\pdfoutput=1

\documentclass[11pt]{article}
\usepackage[table]{xcolor}

\usepackage{ACL2023}
\usepackage[hang,flushmargin]{footmisc}
\usepackage{balance}
\usepackage{changepage}

\usepackage{times}
\usepackage{latexsym}

\usepackage[T1]{fontenc}

\usepackage[utf8]{inputenc}

\usepackage{microtype}

\usepackage{inconsolata}

\usepackage{float}

\usepackage{amsmath}
\usepackage{amssymb}
\usepackage{amsthm}
\usepackage{bm}
\usepackage{booktabs}
\usepackage{graphicx}
\usepackage{multicol}
\usepackage{multirow}
\usepackage{blindtext}
\usepackage{dirtree}
\usepackage{tikz}

\usepackage{caption}
\usepackage{subcaption}
\usepackage{enumitem}
\usepackage{tipa}

\newcommand{\parheader}[1]{{\bf \smallskip \noindent #1.}}
\newcommand{\parheaderfirst}[1]{{\noindent \bf #1.}}

\newcommand{\DONE}[1]{\noindent \textcolor{green}{\textbf{DONE}}\\ }

\DeclareMathOperator*{\argmax}{argmax}

\definecolor{severe}{RGB}{241,113,31}
\definecolor{severe-moderate}{RGB}{215,75,63}
\definecolor{moderate}{RGB}{177,50,90}
\definecolor{moderate-light}{RGB}{135,33,107}
\definecolor{light}{RGB}{92,18,110}

\newtheorem{definition}{Definition}[section]
\newtheorem{proposition}{Proposition}[section]

\title{Words Worth a Thousand Pictures:\ Measuring and Understanding Perceptual Variability in Text-to-Image Generation}

\newcommand{\ignore}[1]{}

\author{Raphael Tang,$^{1,2}$ Xinyu Zhang,$^2$ Lixinyu Xu,$^1$ Yao Lu,$^3$ Wenyan Li,$^4$\\\textbf{Pontus Stenetorp,$^3$ Jimmy Lin,$^2$ Ferhan Ture$^1$} \vspace{1mm}\\
$^1$Comcast AI Technologies~~~$^2$University of Waterloo\\ $^3$University College London~~~$^4$University of Copenhagen\\
{\small $^1$\texttt{{\{firstname\_lastname\}}@comcast.com}~~~$^2$\texttt{\{r33tang, x978zhan, jimmylin\}@uwaterloo.ca}}}

\begin{document}
\maketitle

\begin{abstract}
Diffusion models are the state of the art in text-to-image generation, but their perceptual variability remains understudied.
In this paper, we examine how prompts affect image variability in black-box diffusion-based models.
We propose W1KP, a human-calibrated measure of variability in a set of images, bootstrapped from existing image-pair perceptual distances.
Current datasets do not cover recent diffusion models, thus we curate three test sets for evaluation.
Our best perceptual distance outperforms nine baselines by up to 18 points in accuracy, and our calibration matches graded human judgements 78\% of the time.
Using W1KP, we study prompt reusability and show that Imagen prompts can be reused for 10--50 random seeds before new images become too similar to already generated images, while Stable Diffusion XL and DALL-E 3 can be reused 50--200 times.
Lastly, we analyze 56 linguistic features of real prompts, finding that the prompt's length, CLIP embedding norm, concreteness, and word senses influence variability most.
As far as we are aware, we are the first to analyze diffusion variability from a visuolinguistic perspective.
Our project page is at \url{http://w1kp.com}.
\end{abstract}

\section{Introduction}
\begin{figure}
    \includegraphics[width=\columnwidth]{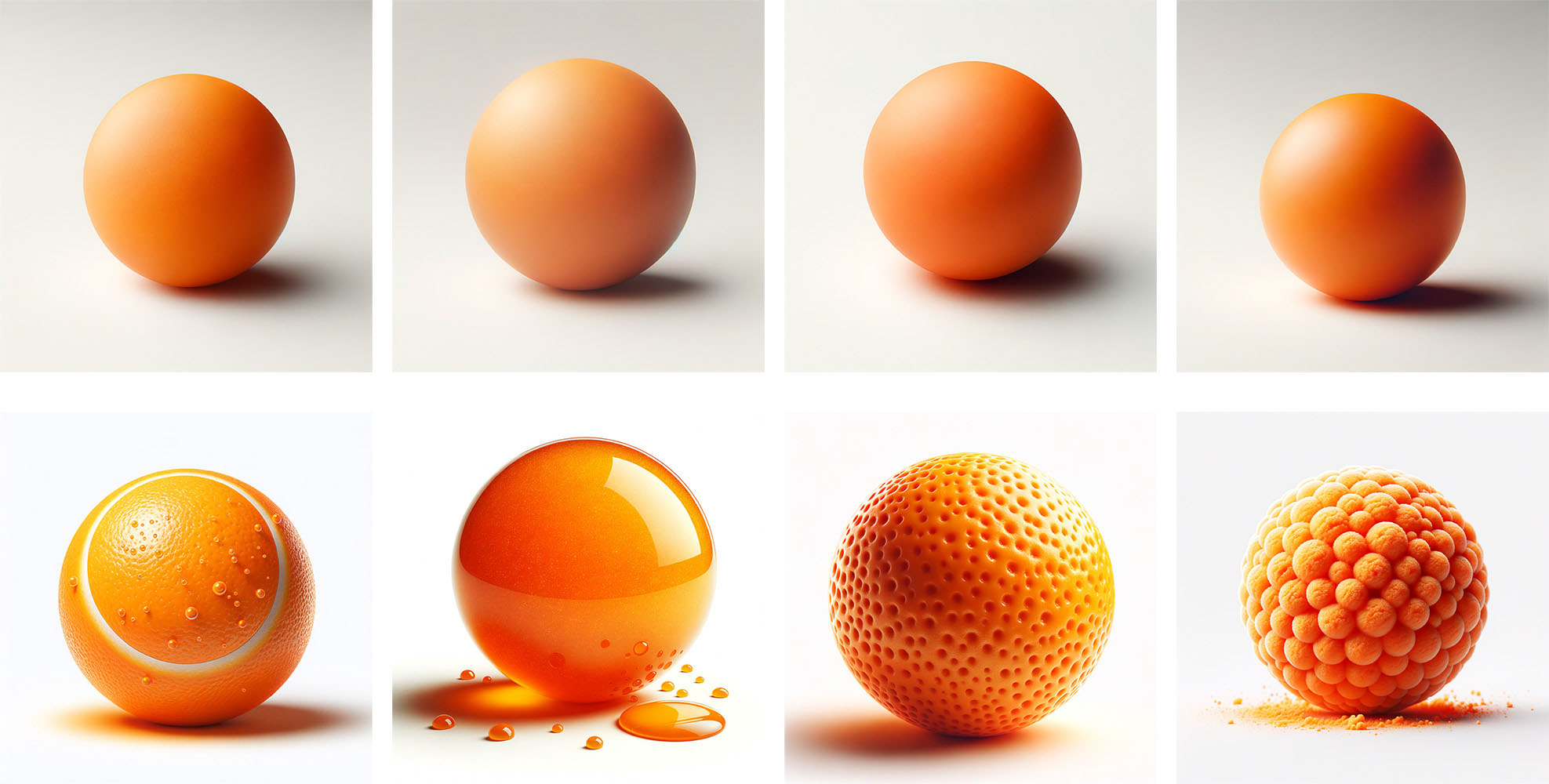}
    \caption{DALL-E 3 images for the prompts ``a matte orange ball in the center against a pure white background''~(top) and ``orange ball against white background''~(bottom).
    Our W1KP score quantifies the perceptual similarity for each set of images.
    It yields 0.99 and 0.68 for the top and bottom rows, showing the greater image variability of the latter.
    }%
    \label{fig:oranges}
\end{figure}

In text-to-image generation, pictures are worth a thousand words, but which words are worth a thousand pictures?
Specifically, how do prompts affect perceptual variation in generated imagery across random seeds?
Consider these prompts:
{\it
    \begin{enumerate}[itemsep=-0.6ex, leftmargin=9mm]\vspace{-1ex}
        \item[\textbf{P1:}] A matte orange ball in the center against a pure white background.
        \item[\textbf{P2:}] Orange ball against white background.
    \end{enumerate}
}\vspace{-1mm}
\noindent As shown in \autoref{fig:oranges}, the first conveys a single particular illustration, while the second elicits multiple interpretations.
Orange could refer to the fruit or the color, and the scene geometry is underspecified. 
But how can we quantify and characterize these linguistic intuitions?

In this paper, we study the connection between visual variability and language in black-box text-to-image models, focusing on state-of-the-art diffusion models.
Previous work tends to study the perceptual distance~\cite{zhang2018unreasonable} between \textit{pairs} of images, while a prompt can generate a near infinite \textit{set} of images.
%
%
Furthermore, previous approaches have not been explicitly calibrated for human-friendly grades of similarity.
What does a score of, for example, 0.2 mean in terms of perceived similarity?
Such calibration is likely crucial for robust human interpretation.

To bridge these gaps in the literature, we first propose a straightforward framework for constructing human-calibrated perceptual \textit{variability} measures based on existing perceptual distance metrics.
We call it the Words of a Thousand Pictures method, or \textit{W1KP}~([\textipa{'wIk.pi:}]) for short.
On our crowd-sourced dataset of human-judged images from DALL-E 3, Imagen, and Stable Diffusion XL~(SDXL), we validate our choice of DreamSim~\cite{fu2024dreamsim}, a recent distance trained on Stable Diffusion~\cite{rombach2022high} images.
Our variant of DreamSim outperforms the best baseline by 0.1--0.4 points in two-alternative forced choice and 0.2--0.4 points in accuracy. 
To improve interpretability, we normalize and calibrate scores to graded human judgements on four levels of perceptual similarity, with cutoff points corresponding to high~(0.85--1.0), medium~(0.4--0.85), low~(0.2--0.4), and no similarity~($<$0.2), which yield a correct classification 78\% of the time.

Next, to ground our academic discourse, we investigate the practical implications of our approach.
Suppose a computer graphics practitioner wishes to generate a diverse array of images from a single prompt, but it is unclear how much it can be reused with different seeds before additional images contribute little to the variability of the overall set of images.
Our work provides a quantitative metric for prompt reusability, as we explore further in Section~\ref{sec:visuolinguistic:prompt-reusability}.
On DiffusionDB~\cite{wang2023diffusiondb}, an open dataset of user-written text-to-image prompts, we find that the same prompt can be reused for Imagen for 10--20 random seeds, while SDXL and DALL-E 3 are more reusable at 100--200 seeds.

Finally, we study how 56 linguistic features affect generation variability.
Although research has explored optimizing for image variability in diffusion~\cite{sadat2024cads}, they have not investigated the contributing linguistic constructs.
To understand the underlying structure of these 56 features, we perform an exploratory factor analysis over DiffusionDB and uncover four factors of keyword presence~(e.g., ``dog walking, \underline{4K}, \underline{watercolor}''), syntactic complexity (e.g., Yngve depth), linguistic unit length, and semantic richness.
Then, we conduct clean-room, single-word generation experiments over the three strongest features in the semantic richness factor~(concreteness, CLIP embedding norm, and number of word senses) to assess their contribution more precisely.
We confirm that all three linguistic features significantly~($p<0.01$) correlate with perceptual variability for all three diffusion models studied.

Our contributions are as follows:\ \textbf{(1)} we propose and validate a human-calibrated framework for building perceptual variability metrics from existing perceptual distance metrics; \textbf{(2)} we examine a new practical application of the method in assessing prompt reusability in text-to-image generation; and \textbf{(3)} we provide original insight into the linguistic sources of variability in diffusion models, finding that keywords, syntactic complexity, length, and semantic richness influence variability.

\section{Our W1KP Approach}

\subsection{Preliminaries}\label{sec:approach:preliminaries}
Text-to-image diffusion models are a family of denoising generative models broadly consisting of two components:\ a text encoder that produces latent representations of language, such as T5~\cite{ raffel2020exploring} or CLIP~\cite{radford2021learning}, and a denoising image decoder that transforms random noise into an image conditioned on text, e.g., a convolutional variational auto-encoder (VAE; \citealp{rombach2022high}).
To generate an image, we feed a prompt into the text encoder, pass its representation to the image decoder along with randomly sampled noise, then iteratively denoise the noise into a meaningful image.
Large-scale models are generally trained using score matching~\cite{song2021scorebased} on billions of image--caption pairs~\cite{podell2024sdxl}, such as the now-deprecated LAION-5B dataset~\cite{schuhmann2022laion}.%

To conduct a general study, we explore diffusion in a black-box manner to be able to generalize to proprietary models.
Formally, let a text-to-image model be $G(\{w_i\}; s, \bm\theta)$ whose codomain comprises the sample space of all images $\mathcal{I}$ and domain the sequence of words $\{w_i\}$, random seed $s \in \mathbb{Z}$ to initialize the image noise, and learned parameters $\bm\theta \in \mathbb{R}^p$.
To generate multiple images from a single prompt, a standard practice is to run multiple trials for different random seeds $s$~\cite{podell2024sdxl}, which we follow in our experiments.

Our analyses target three state-of-the-art models, one open and two proprietary:
\begin{enumerate}[leftmargin=5mm,itemsep=0.25mm,topsep=2mm]
\item \textbf{Stable Diffusion XL}~\cite{podell2024sdxl}, an open model which uses CLIP~\cite{radford2021learning} for encoding text and a 2.6 billion-parameter U-Net~\cite{ronneberger2015u} for generating images.
\item \textbf{DALL-E 3}~\cite{betker2023improving}, a proprietary API from OpenAI incorporating a pretrained T5-XXL~\cite{raffel2020exploring} text encoder and the same image decoder architecture as SDXL.
\item \textbf{Imagen}~\cite{saharia2022photorealistic}, a similarly proprietary API from Google using a T5-XXL encoder and an efficient variant of a similar convolutional U-Net decoder.
\end{enumerate}
All models produce images at least 1024$\times$1024 pixels in resolution.
Further details about the three models can be found in Appendix \ref{appendix:sec:diffusion-model-details}.

\subsection{Our General Framework}\label{sec:approach:framework}

We aim to measure the visual variability of a set of synthetic images.
Toward this, we propose to aggregate perceptual distances, which are well studied in the literature, among all pairs of images in a set.
To aid human interpretation of the distances, we apply two steps:\ first, normalization, which squashes potentially unbounded and ``odd'' distributions into the standard uniform distribution $U[0, 1]$.
For instance, a perceptual distance with a tight range of 5.10--5.19 across 1,000 image sets would be difficult to comprehend.
Second, we calibrate the distances to graded human judgements of similarity and determine the corresponding cutoff points, giving meaning to score ranges~(see \autoref{fig:magnitude-interpretation}).

Concretely, let $\bm I := \{I_i\}_{i=1}^n \subseteq \mathcal{I}$ be an i.i.d.\ sample of images generated by $G(\cdot)$.
We seek a function $\eta(\bm I)$ such that $\eta(\bm I') < \eta(\bm I)$ if $\bm I'$ is more self-similar than $\bm I$ is.
A starting point is perceptual distance, a  symmetric $\delta : \mathcal{I} \times \mathcal{I} \mapsto \mathbb{R}^+$ that assigns larger values to less similar image pairs.
Many metrics~\cite{fu2024dreamsim} embed $I_a, I_b \in \mathcal{I}$ using a feature extractor $f : \mathcal{I} \mapsto \mathbb{R}^\ell$, such as ViT~\cite{dosovitskiy2021an}, then compute a distance $d : \mathbb{R}^\ell \times \mathbb{R}^\ell \mapsto \mathbb{R}^+$ between $f(I_a)$ and $f(I_b)$, e.g., Euclidean distance.
To standardize these distances to $U[0, 1]$ for better interpretability, we apply the cumulative distribution function transform, defined as $F(x) := \mathbb{P}(X \leq x)$.
It has the property of $F(X)$ being uniformly distributed:
\begin{proposition}\label{prop:inverse-cdf}
    If $X$ is a continuous random variable, $F(X)$ is standard uniform $U[0, 1]$.
\end{proposition}
\noindent Hence, a normalized $d^*$ is
\begin{equation}\label{eqn:normalization}
    d^*(I_a, I_b) := F(d(f(I_a), f(I_b))),
\end{equation}
and $F$ is estimated from a sample $\{d(I_{a_i}, I_{b_i})\}_{i=1}^m$ as
$\hat{F}(d(I_a, I_b)) := |\{d(I_{a_i}, I_{b_i}) \leq d(I_a, I_b) : 1 \leq i \leq m\}|/m$.
As our sample, we generate 10,000 image pairs per diffusion model for 1,000 randomly selected DiffusionDB prompts.

Equipped with a uniform perceptual distance, we now construct measures of image set variability~($\eta$).
A natural framework to do this is to define a family of $U$-statistics~\cite{li2012u, hoeffding1948class} over sets of images:
\begin{definition}
     Let $h : \mathbb{R^\ell}\times \cdots \times\mathbb{R}^{\ell} \mapsto \mathbb{R}^+$ be an $\alpha$-arity kernel parameterized by $d$.
     Then a family of $U$-statistics for measuring image set variability can be defined as
    \begin{equation}\small
        U_{d,h}(\bm I) := \frac{1}{\binom{n}{\alpha}}\sum_{1 \leq i_1 < \cdots < i_\alpha \leq n} \hspace{-5mm}h(f(I_{i_1}), \dots, f(I_{i_\alpha}); d).
    \end{equation}
\end{definition}
\noindent Certain choices of $h$ produce estimators of interest.
We use two in our experiments:
\begin{itemize}[leftmargin=4mm,itemsep=0.5mm,topsep=1mm]
    \item \textbf{Pairwise mean} ($\eta_\text{mean})$: let $d=d^*$, $\alpha=2$, and $h(\bm x,\bm y; d) = d(\bm x, \bm y)$. This measures the expected similarity among all pairs of images.
    \item \textbf{$k$-expected maximum} ($\eta_{k}$): let $d=d^*$, $\alpha=k$, and $h(\bm x_1, \dots, \bm x_\alpha) = \min\{d(\bm x_i, \bm x_j) : i\neq j\}$.
    This quantifies the expected maximum similarity between a pair of images in a set of size $k$.
\end{itemize}
\noindent We note a connection to statistical dispersion: if $d$ is the squared Euclidean distance and $h$ the pairwise mean kernel, $U_{d,h}$ is proportional to the trace of the covariance matrix of $f(I_1), \dots, f(I_n)$, i.e., the total variance.
A proof is in Appendix~\ref{appendix:sec:proofs}.
Furthermore, to match the convention of scores in $[0,1]$ denoting similarity rather than dissimilarity (e.g., $R^2$), for the rest of this paper \textit{we invert $\eta$ and report $\tilde{\eta} := 1-\eta$ instead}, calling it the \textit{W1KP score}.

\begin{figure}
    \centering
    \includegraphics[width=\columnwidth,trim={0mm 1mm 0mm 7mm},clip]{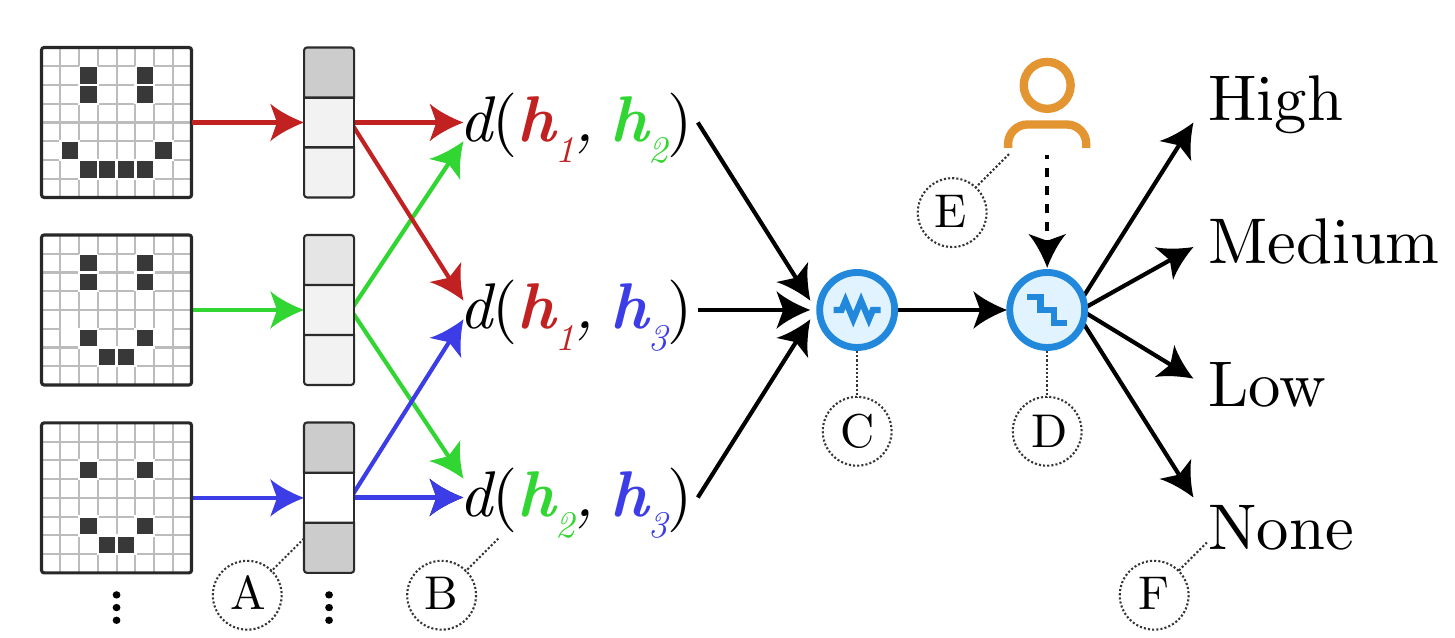}
    \caption{An illustration of W1KP:\ image embeddings~(see A) and pairwise distances (B) computed using a backbone model, fed into the normalization function (C; Eqn.~\ref{eqn:normalization}) producing a single score in $[0, 1]$. The calibration module (D; Eqn.~\ref{eqn:calibration}) aligned to human judgements (E) then assigns a similarity level (F).}
    \label{fig:approach}
\end{figure}

Lastly, we find cutoff points for $\tilde{\eta}$ calibrated to human-judged levels of high, medium, low, and no similarity.
For the human judgement data, we gather a dataset $\{(I_{x_i}, I_{y_i}, z_i)\}_{i=1}^N$, where $I_{x_i}, I_{y_i} \in \mathcal{I}$ are a pair of generated images from the same prompt, and $z_i \in \{\text{none}, \text{low}, \text{mid}, \text{high}\}$ is the human-annotated level of similarity between $I_{x_i}$ and $I_{y_i}$~(see Section~\ref{sec:veracity-analyses:interpretation} for details).
On the dataset, we optimize the cutoff points $\beta_\text{low} < \beta_\text{mid} < \beta_\text{high}$ to maximize the label accuracy of the splits $S_\text{none} := [0, \beta_\text{low})$, $S_\text{low} := [\beta_\text{low}, \beta_\text{mid})$, $S_\text{mid} := [\beta_\text{mid}, \beta_\text{high})$, $S_\text{high} := [\beta_\text{high}, 1.0]$:
\begin{equation}\label{eqn:calibration}\small
    \argmax_{\beta_\text{low}, \beta_\text{mid}, \beta_\text{high}} \frac{1}{N} \sum_{i=1}^N \mathbb{I}(\tilde{\eta}(\{I_{x_i}, I_{y_i}\}) \in S_{z_i}),
\end{equation}
where $\mathbb{I}$ is the indicator function.
We illustrate our overall method in \autoref{fig:approach}, and a proof of Proposition \ref{prop:inverse-cdf} is given in Appendix~\ref{appendix:sec:proofs}.

\section{Veracity Analyses}

\subsection{W1KP Quality}

\begin{table}[t]\small
\setlength{\tabcolsep}{2.2pt}
\centering
\begin{tabular}{lccccccc}
\toprule[1pt]
\multirow{2}{*}{\textbf{Method}\vspace{-2mm}} & \multicolumn{2}{c}{{\textbf{SDXL}}} & \multicolumn{2}{c}{{\textbf{Imagen}}} & \multicolumn{2}{c}{{\textbf{DALL-E 3}}}
\\
\cmidrule(lr){2-3}
\cmidrule(lr){4-5}
\cmidrule(lr){6-7}
& 2AFC & Acc. & 2AFC & Acc. & 2AFC & Acc. \\
\midrule[1pt]
Oracle & 80.0 & 100 & 80.7 & 100 & 79.3 & 100\\
\midrule[0.1pt]
L2 & 54.8 & 55.4 & 61.0 & 63.3 & 58.5 & 60.1\\
SSIM & 55.2 & 56.7 & 59.1 & 61.7 & 57.6 & 59.3\\
\midrule[0.1pt]
LPIPS & 64.7 & 68.6 & 67.6 & 72.0 & 64.8 & 70.8\\
ST-LPIPS & 60.0 & 62.4 & 63.4 & 67.6 & 59.6 & 65.4\\
DISTS & 65.5 & 69.4 & 67.5 & 71.9 & 63.7 & 67.5 \\
SSCD (Large) & 63.4 & 66.7 & 66.0 & 69.1 & 63.3 & 66.7 \\
CoPer (CLIP$_{B32}$) & 63.2 & 67.8 & 64.4& 68.9 & 62.4 & 67.9 \\
Raw (CLIP$_{L14}$) & 67.3 & 72.4 & 70.3 & 76.3 & 67.3 & 75.0\\
\midrule[0.1pt]
DreamSim (Orig.) & 69.2 & 75.0 & 71.3 & 77.3 & 70.3 & 77.9\\
DreamSim$_{\ell_2}$ (Ours) & \textbf{69.3} & \textbf{75.2} & \textbf{71.5} & \textbf{77.5} & \textbf{70.7} & \textbf{78.3}\\
\bottomrule[1pt]
\end{tabular}   
\caption{Quality of the backbones on our evaluation sets, across the image generation model.}
\label{tab:veracity:quality}
\end{table}
Before applying W1KP, we first validate our choice of the perceptual distance backbone.

\parheader{Setup}
Following prior work in perceptual distance evaluation~\cite{zhang2018unreasonable}, we crowd-sourced a dataset of two-alternative forced-choice (2AFC) image triplets using Amazon MTurk~\cite{hauser2016attentive}.
Five unique workers were shown three generated images from the same prompt---a reference image, image A, and image B---and instructed to pick whether A or B resembled the reference more.
This was repeated three times each for 500 random prompts from DiffusionDB, a large dataset of user-written prompts, for each of SDXL, Imagen, and \mbox{DALL-E 3}, totaling 1,500 triplets per model.
Formally, let $\{(I_{r_i}, I_{a_i}, I_{b_i}, y_{a_i})\}_{i=1}^M$ be a dataset of $M$ triplets, where $I_{r_i}, I_{a_i}, I_{b_i} \in \mathcal{I}$ are images and $y_{a_i} \in \{0, \dots, 5\}$ the number of workers choosing $I_{a_i}$ over $I_{b_i}$.
We used attention checks throughout the process; for more details, see Appendix~\ref{appendix:sec:annotation-apparatuses}.

For our non-neural methods, we evaluated raw-image Euclidean distance (L2) and the structural similarity index~(SSIM; \citealp{wang2004image}).
For our neural backbones, we tested the popular LPIPS~\cite{zhang2018unreasonable}, its shift-tolerant variant ST-LPIPS~\cite{ghildyal2022shift}, and an SSIM-inspired variant DISTS~\cite{ding2020image}, all based on the VGG-16 architecture~\cite{simonyan2015very}; SSCD~\cite{pizzi2022self}, a model trained for image copy detection; CoPer~\cite{li2022domain}, an extension of LPIPS to ViT; raw cosine similarity scores from CLIP~\cite{radford2019language}; and lastly, DreamSim~\cite{fu2024dreamsim}, which ensembles pretrained transformers trained on Stable Diffusion images for feature extraction and applies cosine distance for measurement.
Since DreamSim's domain was closest to ours, we hypothesized that it would be most effective.
We also evaluated our variant, DreamSim$_{\ell_2}$, with Euclidean instead of cosine distance for $d$, which benefits from being a true mathematical distance and hence allows for multidimensional scaling analyses, as in Appendix~\ref{appendix:sec:mds}.

We used the standard evaluation metrics of 2AFC score, defined as the mean proportion of workers agreeing with the backbone's scores, i.e., $\frac{1}{M}\sum_{i=1}^M \mathbb{I}(I_{a_i}\succ_{\hspace{-0.5mm}r} I_{b_i})\frac{y_{a_i}}{5} + \mathbb{I}(I_{a_i}\prec_{\hspace{-0.2mm}r} I_{b_i})(1 - \frac{y_{a_i}}{5})$, where $I_{a_i} \prec_{\hspace{-0.2mm}r} I_{b_i}$ if $\tilde{\eta}(\{I_{r_i}, I_{a_i}\}) < \tilde{\eta}(\{I_{r_i}, I_{b_i}\})$, and majority-vote accuracy.
We let $\tilde{\eta} = \tilde{\eta}_\text{mean}$. 
See Appendix~\ref{appendix:sec:annotation-apparatuses} for further setup details.

\parheader{Results}
We present our results in \autoref{tab:veracity:quality}.
As an upper bound, we report the maximum possible 2AFC and accuracy in row one.
In line with intuition, our DreamSim backbones attain the highest quality, surpassing CLIP$_{L14}$ raw, the second best, by 2.0 points in 2AFC and 2.8 in accuracy on average.
Our variant DreamSim$_{\ell_2}$ slightly outperforms the original DreamSim with statistical significance~($p<0.05$ on the paired $t$-test) by 0.1--0.4 in 2AFC and 0.2--0.4 in accuracy, possibly since the embedding norm is informative~\cite{oyama2023norm}.
Thus, we select DreamSim$_{\ell_2}$ as the backbone for W1KP.

Beyond quality assurance, another purpose of this evaluation is to ensure that the backbone does equally well on the three image generators.
As a sanity check, the oracle~(row one) has a spread of 1.4 points~(79.3--80.7) in 2AFC on the three models, indicating that humans are unbiased.
Our DreamSim$_{\ell_2}$ has a spread of 2.2 points~(69.3--71.5) in 2AFC, which is below the global average spread of 3.3 points for all the methods.
We conclude that DreamSim$_{\ell_2}$ exhibits less model-wise bias than its counterparts, possibly due to its increased quality and in-domain training.

A potential issue is that perceptual similarity is inherently subjective and hence challenging to measure.
Research suggests to also evaluate just-noticeable differences~(JND), which is thought to be cognitively impenetrable due to its viewing-time constraint~\cite{acuna2015using}.
Because of the high correlation between 2AFC and JND on synthetic images ($r=0.94$; \citealp{fu2024dreamsim}), 2AFC appears to be a viable proxy for JND for our study.

\subsection{W1KP Metric Interpretation}\label{sec:veracity-analyses:interpretation}

We now assess the quality of our human calibration process, as described near the end of Section~\ref{sec:approach:framework}.

\parheader{Setup}
We collected a dataset of graded image pairs with MTurk.
For 500 random DiffusionDB prompts, three unique workers were presented with two images generated from the same prompt and asked to judge the similarity on an integral scale ranging from ``not similar at all'' (rating 1) to ``the same'' (5).
Afterwards, we merged the last two categories (``same'' and ``very similar'') since the fifth was mostly reserved for attention checks, resulting in the final four categories of high, medium, low, and no similarity.
We took the median across the three judgements and repeated the process for SDXL, Imagen, and \mbox{DALL-E 3}, for a total of 1,500 median judgements roughly split into 10\%, 30\%, 40\%, and 20\% for ratings 1--4.
Our evaluation then consisted of applying Eqn.~\eqref{eqn:calibration} with five-fold cross validation.
For detailed settings, see Appendix~\ref{appendix:sec:annotation-apparatuses}.

\begin{figure}[t]
    \centering
    \includegraphics[width=0.986\columnwidth]{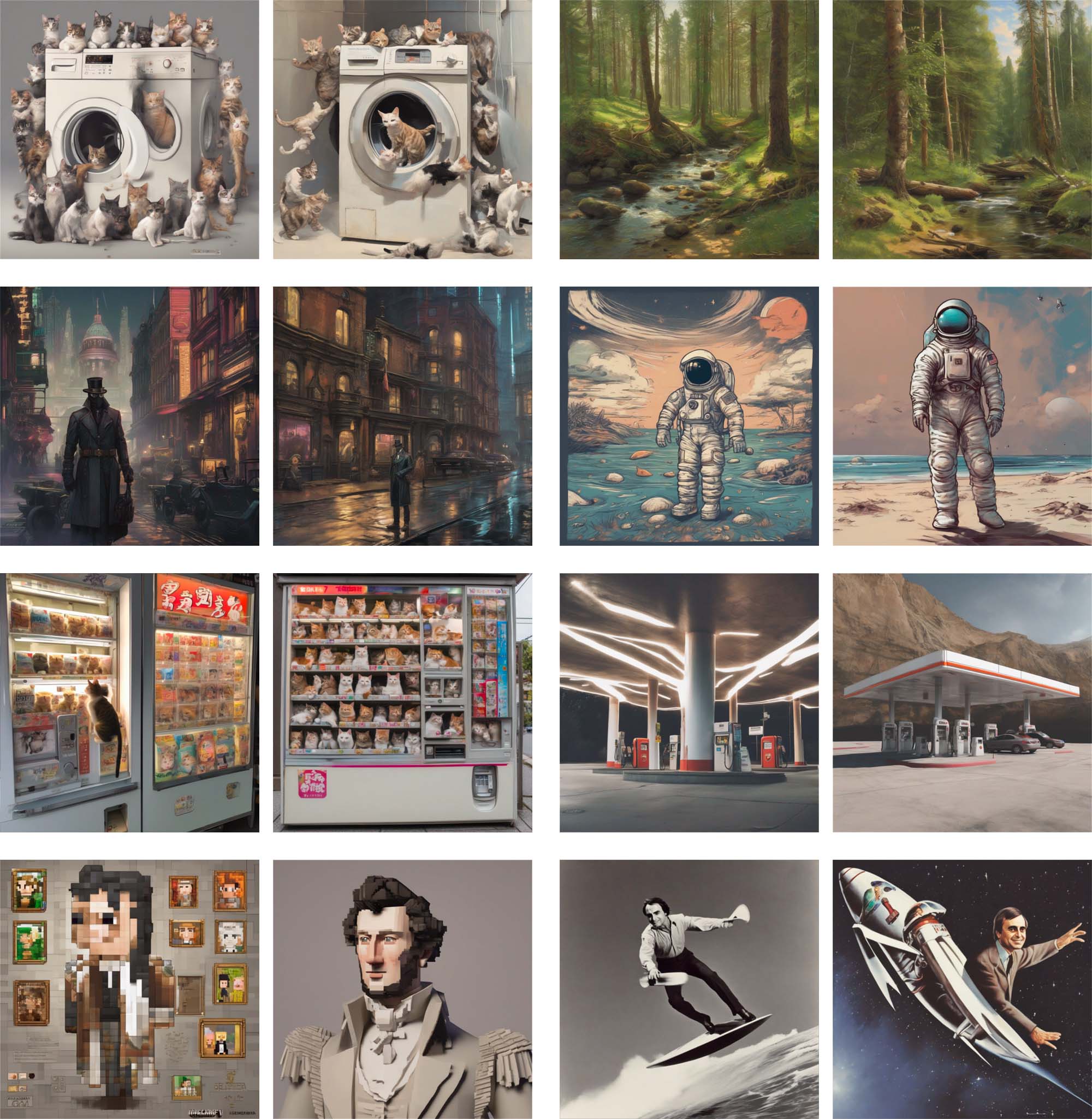}
    \caption{Image pairs from SDXL, ordered row-wise by calibrated W1KP scores. From top to bottom, the rows correspond to high (0.85--1.0), medium (0.4--0.85), low (0.2--0.4), and no similarity (0.0--0.2).}\label{fig:magnitude-interpretation}
\end{figure}

\parheader{Results}
Eqn.~\eqref{eqn:calibration} yields cutoff points (rounded to the nearest $0.05$ for memorability) of $0.2$, $0.4$, and $0.85$  for $\beta_\text{low}$, $\beta_\text{mid}$ and $\beta_\text{high}$.
Overall, we attain macro- and micro-accuracy scores of 80\% and 78\% with DreamSim$_{\ell_2}$ as the backbone.
For comparison, the average macro-/micro-accuracy scores of humans are 82\%/80\%.
DreamSim$_{\ell_2}$ also outperforms the original DreamSim, which has a macro-/micro-accuracy of 79\%/77\%.
Thus, we conclude that our calibration yields interpretable cutoffs.

We present qualitative examples of our cutoffs in \autoref{fig:magnitude-interpretation}.
The levels appear sensible: ``high'' pairs (top row) match in low-level features (e.g., trees in the same location), high-level composition (e.g., cats in washing machine), artistic style (e.g., color photography); medium (second) in composition and style; low (third) in style; and none (last) mostly differing in all.
This aligns with our quantitative results in Appendix~\ref{appendix:sec:metric-interpretation}.
We also verify that normalization (Eqn.~\ref{eqn:normalization}) is necessary.
Before normalization, raw W1KP scores have 10$^\text{th}$, 50$^\text{th}$, and 90$^\text{th}$ percentiles of 0.4, 0.7, and 1.1, which is significantly nonuniform ($p < 0.01$; KS test).

\begin{figure}[t]
\includegraphics[width=\columnwidth]{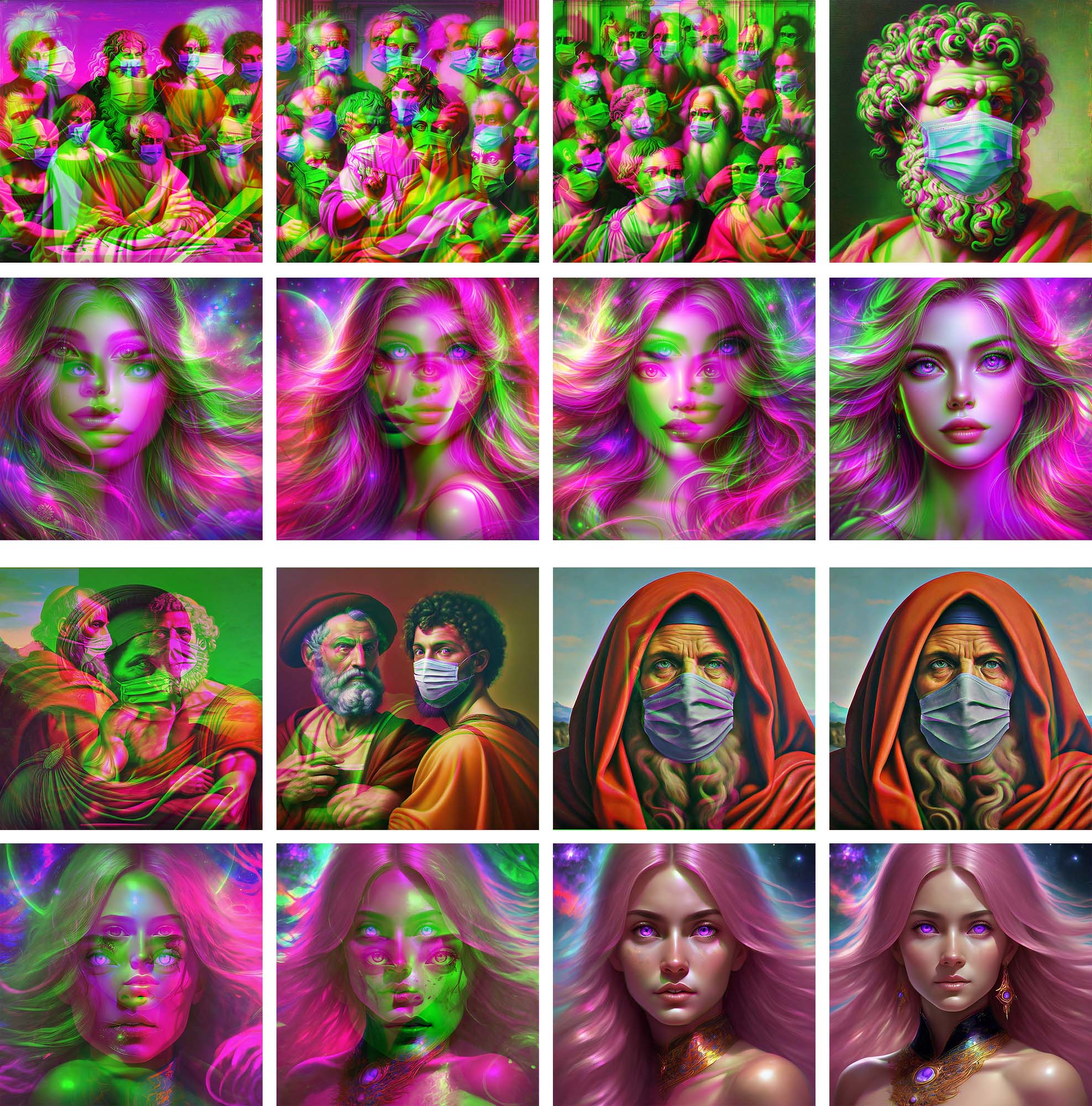}
\caption{Visualizing the overlap between the two most similar images~(on average) as we generate more images for the two prompts. We remove the green channel for one image (magenta) and keep only the green for the other, then stack the two. Above, Imagen is reusable up to 10--50 images, while \mbox{DALL-E 3} up to 50--200.}\label{fig:reusability-visualization}
\end{figure}

One conceivable question is whether calibration and normalization are essential for downstream analysis.
It can be argued that analytic conclusions may still hold without a normalized, calibrated metric.
However, as alluded to in Section~\ref{sec:approach:framework}, there are two clear benefits to having one: first, normalization scales arbitrary scores to the 0--1 range, in line with other common statistics such as $F_1$ score and $R^2$.
Our normalized score also has the direct interpretation as the percentile of the raw score on a known ground-truth distribution.
Second, calibration allows us to interpret scores and aid human understanding.
In Section~\ref{sec:visuolinguistic:prompt-reusability} for example, we use $\beta_\text{high}$ as a cutoff for prompt reusability.

\section{Visuolinguistic Analyses}
With the variability metric established, we investigate the connection between visual variability and prompt language for text-to-image models.

\subsection{Prompt Reusability Analysis}
\label{sec:visuolinguistic:prompt-reusability}

We first ask how many times a prompt can be reused (under different random seeds) until new images are too similar to already generated ones.
This applies to graphic asset creation in particular, where visual artists are tasked with rendering many images of the same concept.
To study this quantitatively, we sampled 50 random prompts from DiffusionDB, generated 300 images for each prompt using different seeds on SDXL, Imagen, and DALL-E 3, then computed the $k$-expected maximum $\tilde{\eta}_k$ for $k=1, \dots, 300$.

\begin{figure}[t]
\centering
\includegraphics[width=\columnwidth,trim={0cm 0.8cm 0cm 0.6cm},clip]{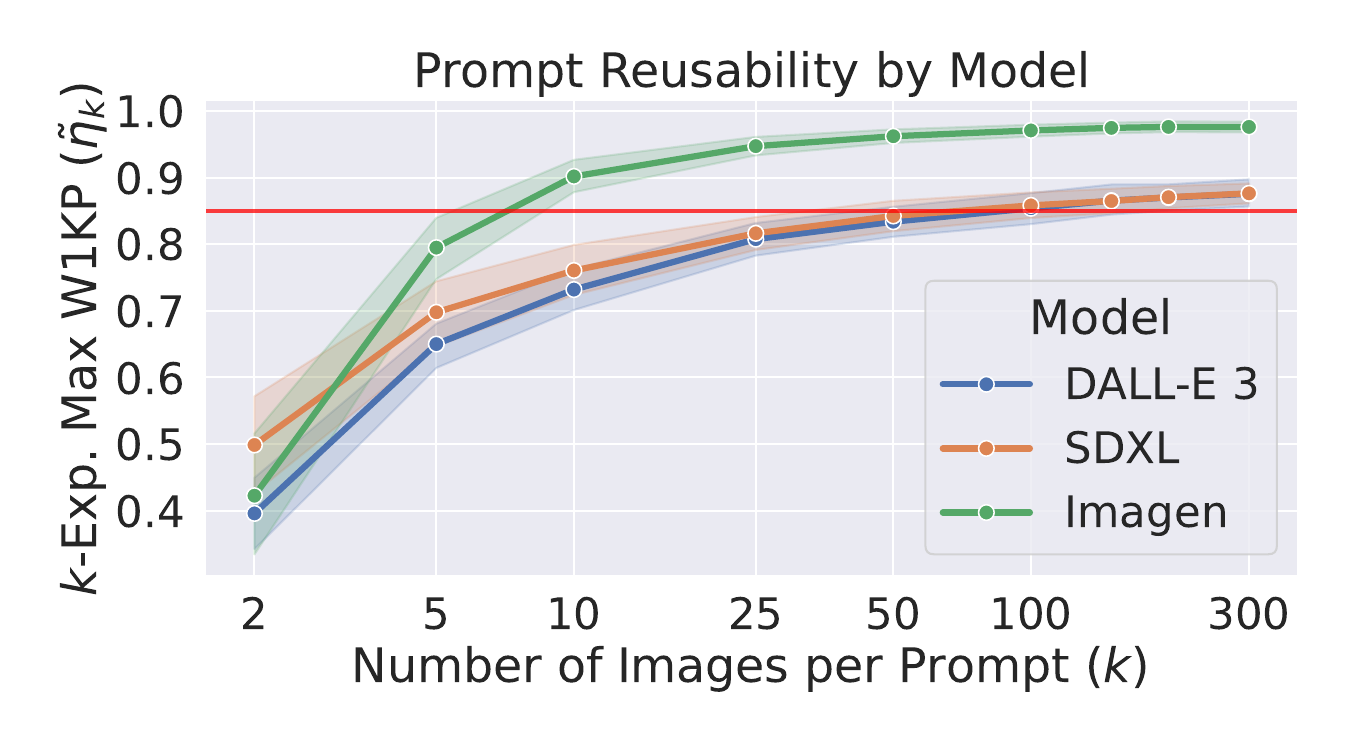}
\caption{$k$-expected maximum ($\tilde{\eta}_k$) for $k=2$ to $300$. Shaded regions denote 95\% confidence intervals and the red line $\beta_\text{high}$.}\label{fig:reusability-plot}
\end{figure}

As visualized in \autoref{fig:reusability-visualization} and plotted in \autoref{fig:reusability-plot}, our diffusion models vary in reusability.
DALL-E 3 on average does not generate highly similar images ($\tilde{\eta}_k \geq \beta_\text{high})$ until $k \to 200$, with our visualization (top two rows in \autoref{fig:reusability-visualization}, one prompt each) displaying much green- and magenta-shifting until the last column.
On the other hand, Imagen tends to produce duplicate images for $k \to 50$.
At 50 images, the two overlaid images are nearly indistinguishable from the true-color image; see the third column.
\autoref{fig:reusability-plot} corroborates these visual results, with the red line ($\beta_\text{high}$) intersecting Imagen's green line between 5--10 and DALL-E 3's blue line at 50--100.
It also suggests that SDXL resembles DALL-E 3 in prompt reusability; see the overlap between the two.
We conclude that diffusion models differ in prompt reusability, possibly due to different decoder architectures.
For example, DALL-E 3 and SDXL share the same U-Net architecture, whereas Imagen's is sparsified~\cite{saharia2022photorealistic}.

\subsection{Exploratory Factor Analysis}
\label{sec:visuolinguistic:factor-analysis}

Our next two analyses relate various linguistic features of prompts such as syntactic complexity to perceptual variability.
First, to understand the salient structure of these linguistic features, we conduct a factor analysis over DiffusionDB.

\parheader{Setup}
Our analysis emulates previous work in interpreting linguistic features for speech~\cite{fraser2016linguistic}.
We extracted 56 features for each of the 1,000 random prompts:
\begin{itemize}[leftmargin=4mm,itemsep=-0.5mm,topsep=1mm]
\item \textbf{Syntactic complexity}: 24 scalar features related to syntax comprehension, such as clauses per T-unit and mean T-unit length, extracted using L2SCA~\cite{lu2010automatic}. We also added Yngve depth, a measure of embeddedness~\cite{yngve1960model}. Our motivation was that sentences with more qualifiers and nominals may be more visually precise.
\item \textbf{Keywords}: 20 Boolean features indicating the presence of the top-20 keywords. We had noticed that most prompts contained trailing keyword qualifiers after a noun phrase, e.g., ``cat beside road, \underline{4k}'' (see Appendix \ref{appendix:sec:diffusiondb} for more); thus, we extracted the top 20 as features.
\item \textbf{Word order}: 3 Boolean features denoting the presence of the PTB~\cite{marcinkiewicz1994building} part-of-speech patterns ``NN VB,'' ``NN VB RB,'' and ``JJ NN'' in the prompt. Our purpose was to assess the effects of adjectives and verbs on nouns.
\item \textbf{Psycholinguistics}: 4 features in mean concreteness judgements \cite{brysbaert2014concreteness}, richness (Honore's statistic and whether a word was in a 100k-word dictionary), and word frequency~\cite{brysbaert2009moving}. 
\item \textbf{Semantic relations}: 3 scalars for the mean number of hyponyms, hypernyms, and word senses, from WordNet~\cite{miller1995wordnet} enhanced with word sense clustering~\cite{snow2007learning}. Intuitively, words with many synonyms (e.g., ``saw'') or hyponyms (e.g., ``animal'') may have more visual representations.
\item \textbf{Embedding norm}: 2 scalars for the mean square GloVe norm~\cite{Pennington2014GloVeGV} and CLIP embedding norm~\cite{radford2021learning}. Word embedding norms were found to encode information gain~\cite{oyama2023norm}, which may affect perceptual variability through specificity.
\end{itemize}

\noindent We generated 20 images per prompt for SDXL, Imagen, and DALL-E 3 and used
Stanford CoreNLP~\cite{manning2014stanford} as our parser~(additional details in Appendix~\ref{appendix:sec:linguistic-feature-extraction}).

\begin{table}[t]\small
    \setlength{\tabcolsep}{2.2pt}
    \centering
    \resizebox{\columnwidth}{!}{%
    \begin{tabular}{rlcccc|cc}
\toprule[1pt]
\# & \textbf{Name} & \textbf{Fac. 1} & \textbf{Fac. 2} & \textbf{Fac. 3} & \textbf{Fac. 4} & $\rho$ & $\mu$ \\ 
\midrule[1pt]\\[-2.85ex]
\multicolumn{8}{c}{Factor 1:\ Style Keyword Presence; Mean $|\rho|=0.12$\vspace{-0.75mm}}\\
\midrule[0.5pt]
1  & Keyword:\ cgsociety & \textbf{0.80} &          &           &           & 0.09 & 0.05\\
2  & Keyword:\ 8k        & \textbf{0.75} &          & -0.10     &           & 0.12 & 0.17\\
3  & Keyword:\ detailed  & \textbf{0.75} &          &           &           & 0.14 & 0.05\\
4  & Keyword:\ artgerm   & \textbf{0.66} &          &           &           & 0.15 & 0.06\\
5  & Keyword:\ cinematic & \textbf{0.59} &          &           &           & 0.11 & 0.04\\
6  & Keyword:\ digital art & \textbf{0.43} &          &           &           & 0.10 & 0.04\\
\midrule[1pt]\\[-2.85ex]
\multicolumn{8}{c}{Factor 2:\ Syntactic Complexity; Mean $|\rho|=0.09$\vspace{-0.75mm}}\\
\midrule[0.5pt]
7  & Clauses per T-unit (T)           &          & \textbf{1.08} & -0.13     & -0.13     & 0.07 & 0.69\\
8  & Clauses per sentence         &          & \textbf{0.92} &           & -0.13     & 0.05 & 0.69 \\
9 & Number of T-units           &          & \textbf{0.63} & \textbf{-0.37} & 0.20  & 0.07 & 0.92 \\
10 & Verb phrases/T      & -0.11    & \textbf{0.47} &           &           & 0.05 & 0.50\\
11 & Complex nominals/T  & -0.12    & \textbf{0.46} & \textbf{0.46}  & 0.12  & 0.19 & 2.16 \\
\midrule[1pt]\\[-2.85ex]
\multicolumn{8}{c}{Factor 3:\ Linguistic Unit Length; Mean $|\rho|=0.19$\vspace{-0.75mm}}\\
\midrule[0.5pt]
12 & Mean T-unit length        &          & \textbf{0.49} & \textbf{0.60}  & 0.17  & 0.18 & 16.7 \\
13 & Mean clause length        &          & \textbf{0.45} & \textbf{0.53}  & 0.19  & 0.18 & 15.9 \\
14 & Mean sentence length      &          &          & \textbf{0.51}  & \textbf{0.45}  & 0.27 & 21.4 \\
15 & Coordinate phrases/T & 0.15     & 0.20     & 0.27      &           & 0.13 & 0.33\\
\midrule[1pt]\\[-2.85ex]
\multicolumn{8}{c}{Factor 4:\ Semantic Richness; Mean $|\rho|=0.17$\vspace{-0.75mm}}\\
\midrule[0.5pt]
16 & Number of words              &          & 0.12     & 0.11      & \textbf{0.75}  & 0.30 & 24.6\\
17 & CLIP embedding norm          &          &          & 0.17      & \textbf{-0.61} & -0.31 & 151 \\
18 & \texttt{ADJ NOUN} & & & & \textbf{0.55} & 0.21 & 0.82\\
19 & Percentage of keywords       & 0.20     & 0.11     &           & \textbf{0.55}  & 0.20 & 48.8 \\
20 & Mean concreteness            &          &          &           & \textbf{0.47}  & 0.25 & 2.30 \\
21 & Mean \# of word senses   & -0.11    &          &           & \textbf{0.43}  & -0.18 & 2.58 \\
22 & Honore's statistic           &          &          &           & \textbf{-0.38} & -0.09 & 7.36 \\
23 & Not in dictionary            &          &          &           & 0.29      & 0.09 & 0.91\\
24 & Keyword:\ elegant   &          &          &           & 0.21      & 0.04 & 0.04 \\
25 & Keyword:\ fantasy &      &          &           & 0.15      & 0.05 & 0.04 \\
\bottomrule[1pt]
\end{tabular}}
    \caption{Linguistic features grouped by interpreted factors, with high loadings ($\geq$0.3) in bold and low loadings~($<$0.1) removed. All Spearman's $\rho$ are statistically significant ($p<0.05$); insignificant features omitted.}
    \label{tab:results}
\end{table}

\parheaderfirst{Results}
We present our results in \autoref{tab:results}.
Following standard practice~\cite{fraser2016linguistic}, we use an oblique promax rotation to enable interfactor correlation.
Four factors capture sufficient variance according to Kaiser's criterion~\cite{kaiser1958varimax}.
For each feature, we report its correlation (Spearman's $\rho$) with the per-prompt perceptual similarity ($\tilde{\eta}_\text{mean}$) and compute the mean feature score $\mu$.

As is conventional, we manually explain the four factors (F1--F4).
For F1, ``8k,'' ``detailed,'' ``cinematic,'' and ``digital art'' describe the art style, ``cgsociety'' pertains to computer graphics, and ``artgerm'' is an artist with a specific style; hence, we call it ``style keyword presence.''
F2's features are classic measures of syntactic complexity~\cite{lu2010automatic} and thus labeled as such.
In F3, mean length of clauses, sentences, and T-units quantify various lengths, so we name it ``linguistic unit length.''
Lastly, F4 primarily depicts semantic richness, with
concreteness, CLIP embedding norm (related to information gain), number of word senses, and \texttt{ADJ NOUN} roughly characterizing visual (non)ambiguity and Honore's statistic, the number of words, and ``not in dictionary'' portraying lexical richness.

Our feature correlations with W1KP agree with intuition.
Having higher concreteness (e.g., house vs.\ dignity) and fewer word senses (saw vs.\ tomato) increases similarity (rows 20, 21), likely since abstract and polysemous words have more visual interpretations. 
Complex nominals (row 11), adjectival modifiers (row 18), and keywords (F1) limit variability through qualification.
Semantic richness has the strongest correlated features, with half having $|\rho|\hspace{-0.8mm}>\hspace{-0.8mm}0.2$.
CLIP norm is the most predictive of variability ($\rho\hspace{-0.8mm}=\hspace{-0.8mm}-0.31$), possibly because text embeddings from vision-language models are used to initialize image generation (Sec.\ \ref{sec:approach:preliminaries}).
Larger norms may yield more chaotic decoding trajectories in the iterative solver, increasing variability.
Factor-wise, linguistic unit length has the highest mean $|\rho|$ of $0.19$, where sentence length is the third most predictive feature ($\rho\hspace{-0.8mm}=\hspace{-0.8mm}0.27$).
Longer prompts presumably provide more visual information.
We conclude that many features in the linguistic space are predictive of variability in the visual space, especially CLIP norm, length, and concreteness.

\subsection{Confirmatory Lexical Analysis}
The last section studies how prompts relate to variability in the DiffusionDB corpus.
While it benefits from realism, some experimental control is lost.
Thus, to supplement the previous study, this section uses single-word synthetic prompts, sampled and adjusted for word frequency in a clean-room manner.
We examine the effects of concreteness, CLIP norm, and polysemy---three of the strongest features from Sec.\ \ref{sec:visuolinguistic:factor-analysis}.

\parheader{Setup}
For our prompts, we sampled 500 words from the 10k most common words in the Google Trillion Word Corpus~\cite{brants2006web}.
We noted each word's concreteness rating ($x_\text{conc})$, number of word senses ($x_\text{sens}$), CLIP embedding norm ($x_\text{clip}$), and frequency rank ($x_\text{freq}$) as our explanatory variables, mirroring the setup of Section~\ref{sec:visuolinguistic:factor-analysis}.
Words without concreteness ratings were resampled.
We then generated 20 images for each prompt with SDXL, Imagen, and \mbox{DALL-E 3} and measured perceptual variability using $\tilde{\eta}_\text{mean}$.
For our analysis, we fit a linear mixed model with $x_\text{conc}$, $x_\text{sens}$, $x_\text{clip}$, and $x_\text{freq}$ as the fixed effects, an intercept for each diffusion model as the random effect, and $\tilde{\eta}_\text{mean}$ as the response variable.
Our purpose is to test whether concreteness, polysemy, CLIP norm, and word frequency independently influence perceptual variability for each model. 

\begin{figure}
    \centering
    \includegraphics[width=0.49\columnwidth,trim={0.7cm 0.7cm 0.05cm 0.7cm},clip]{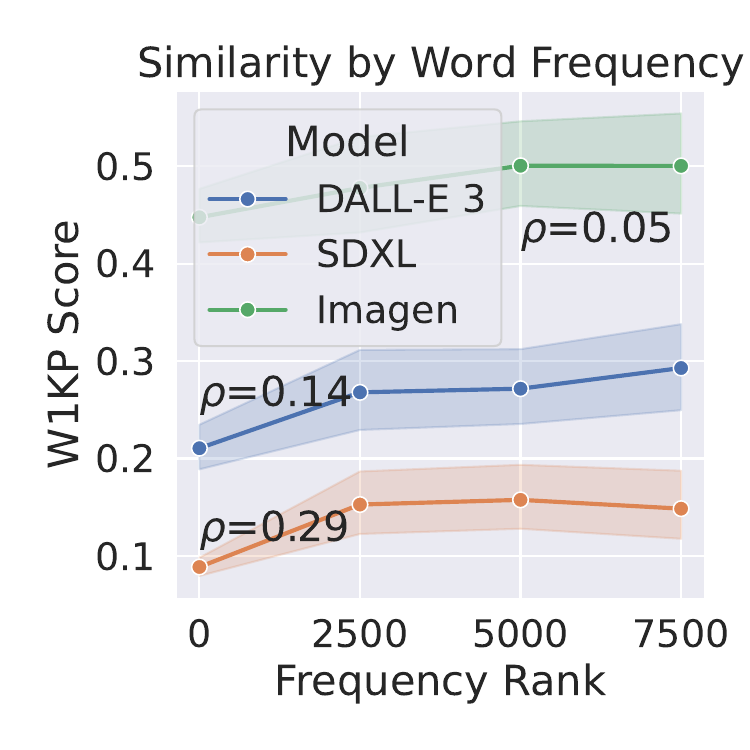}
    \includegraphics[width=0.49\columnwidth,trim={0.7cm 0.7cm 0.05cm 0.7cm},clip]{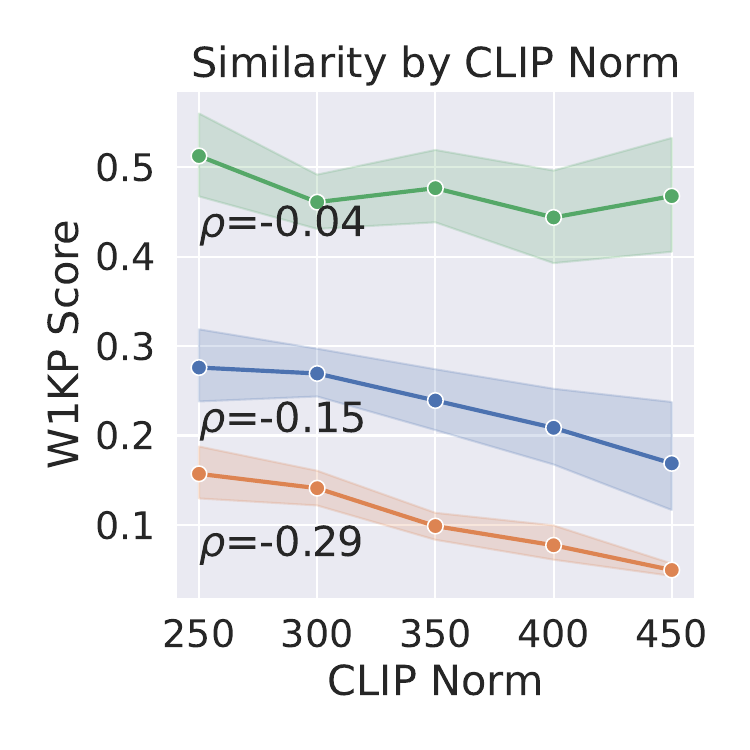}
    \\[1ex]
    \includegraphics[width=0.49\columnwidth,trim={0.7cm 0.7cm 0.05cm 0.7cm},clip]{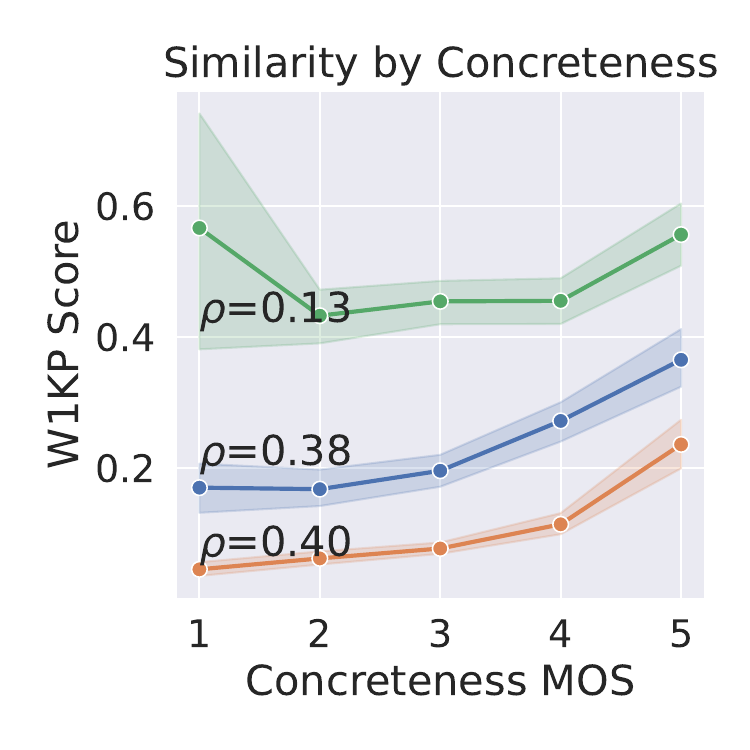}
    \includegraphics[width=0.49\columnwidth,trim={0.7cm 0.7cm 0.05cm 0.7cm},clip]{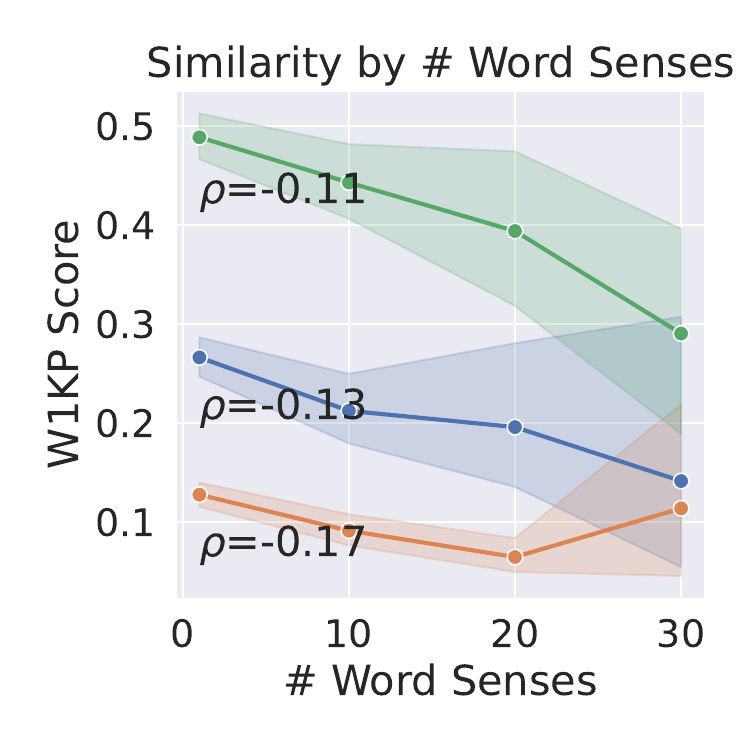}
    \caption{A plot of $\tilde{\eta}_\text{mean}$ against frequency, CLIP norm, concreteness, and word senses for single-word prompts. Shaded regions are 95\% confidence intervals.}\label{fig:lex-analysis}
\end{figure}

\parheaderfirst{Results}
Our linear mixed model reveals statistically significant relationships ($p\hspace{-0.7mm}<\hspace{-0.7mm}0.01$) between $\tilde{\eta}_\text{mean}$ and all the predictors, whose coefficients are $2.4\times10^{-3}$, $4.7\times10^{-4}$, $-7.8\times10^{-5}$, and $-7.2\times10^{-2}$ for $x_\text{sens}$, $x_\text{clip}$, $x_\text{freq}$, and $x_\text{conc}$, respectively.
In other words, polysemy, CLIP norm, word frequency, and concreteness are significant independent factors for perceptual variability, where polysemy and CLIP norm are positively correlated, while frequency and concreteness negatively so.
In \autoref{fig:lex-analysis}, our feature-wise plots further illustrate each individual fixed effect.
The correlation scores are consistent in direction across the diffusion models, with similar signs in Spearman's $\rho$ for each feature.
They also differ by an additive shift, affirming our random-intercepts mixed model.

\begin{figure}[t]
\includegraphics[width=\columnwidth]{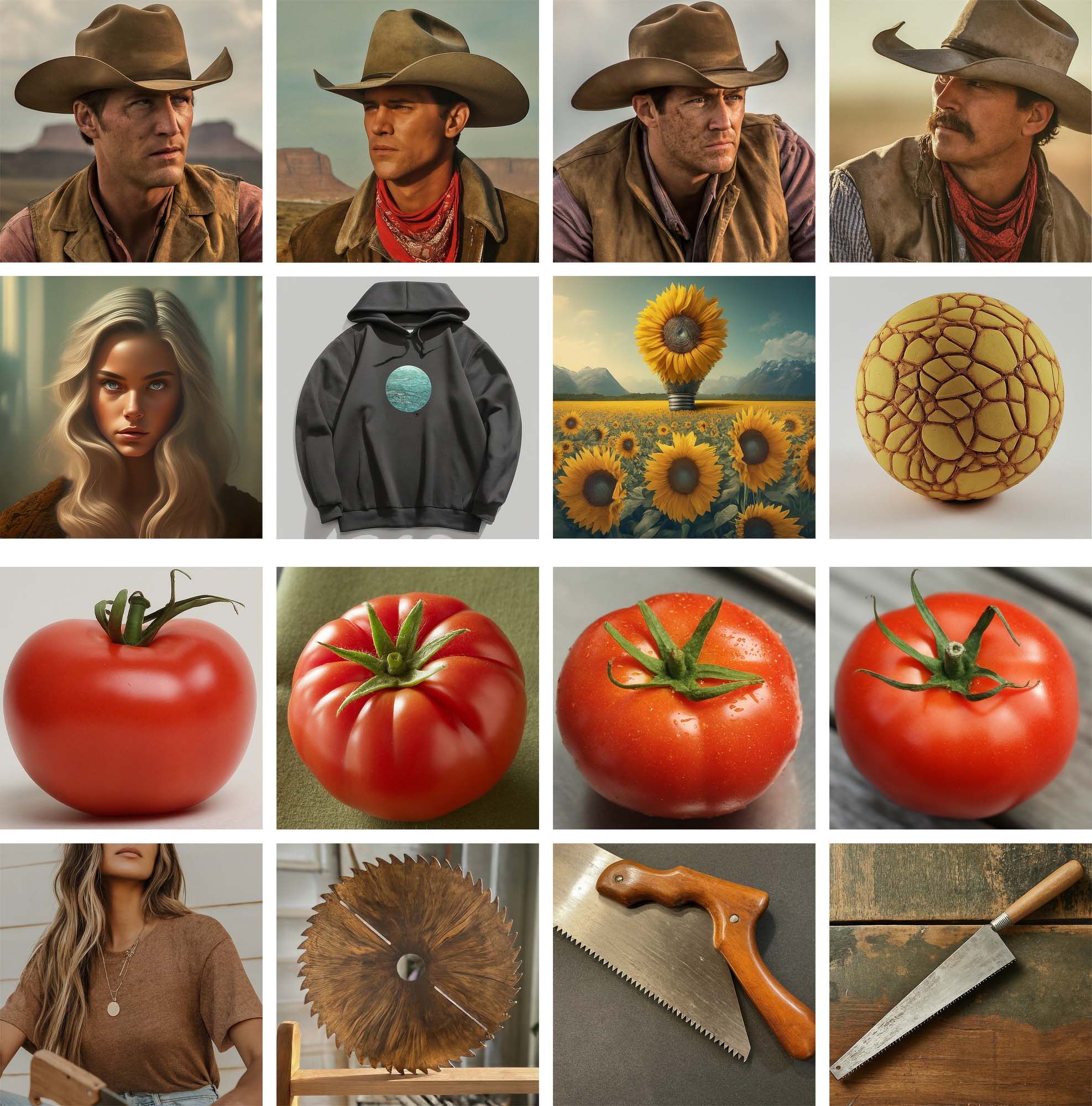}
    \caption{Four single-word Imagen prompts with varying concreteness (``cowboy'' vs. ``concept'') and number of word senses (``tomato'' vs. ``saw'').}\label{fig:lex-examples}
\end{figure}

\autoref{fig:lex-examples} presents prompts of varying concreteness and senses.
``Cowboy,'' a concrete prompt, is less variable than ``concept,'' an abstract one, since a cowboy is tangible.
``Tomato,'' a monosemous word, has less variability than ``saw,'' a polysemous word, because it has a narrow visual representation.
In summary, our exploratory findings on concreteness, CLIP norm, and polysemy from Section~\ref{sec:visuolinguistic:factor-analysis} hold in the clean-room single-word prompt setting.

\section{Related Work and Future Directions}
A related line of work examines boosting image variability in diffusion models~\cite{zameshina2023diverse, sadat2024cads, gu2024kaleido}.
Complementary to their work, our paper analyzes the precise linguistic features contributing to variability.
One future direction could be to incorporate these features into the optimization of variability.

Previous work has analyzed diffusion models using a mixture of computational linguistics and vision techniques.
\citet{tang2023daam} conducted an attribution analysis over Stable Diffusion and discovered entanglement, to which \citet{rassin2024linguistic} proposed to fix using attention alignment. 
Separately, \citet{toker2024diffusion} studied the layer-intermediate representations of diffusion, showing that rare concepts require more computation.
A further extension could be to study linguistic features responsible for increased computation, as our paper also relates word rarity to variability.

Finally, research has previously scrutinized the~(lack of) variability in older architectures such as VAEs~\cite{razavi2019generating} and generative adversarial networks, e.g., mode collapse.
In this paper, we extend this analysis to modern diffusion models while taking a visuolinguistic perspective.

\section{Conclusions}
In conclusion, we examined the connection between visual variability and prompt language for black-box diffusion models.
We proposed a framework for quantifying and calibrating visual variability, applying it to study prompt reusability and linguistic feature salience.
After validating it quantitatively, we found that length, embedding norm, and concreteness influence variability the most.

\section*{Limitations}
One limitation of our work is that while we analyzed the inference-time behavior of various diffusion models, we did not trace the training-time cause of perceptual variability due to the scope of our study.
Doing so would require the training of multiple diffusion models while varying the training sets, which is beyond our budget.

Another limitation is that we have not meticulously characterized the precise distribution of perceptual variability relative to various levels of linguistic features, with our analyses constrained to averages due to the moderate sample size.
For instance, does Imagen yield a higher maximum variability for certain levels of concreteness, even if on average it is lower?
Are there subgroups within each feature that better explain variances in perceptual variability?
Such questions require a larger sample size to answer.

We also consciously limited our examination to random seeds and dispensed with comprehensively assessing other factors possibly influencing perceptual variability, such as classifier-free guidance~\cite{ho2021classifier}.
We vary the guidance scale in Appendix~\ref{appendix:sec:cfg} to confirm that SDXL is always more diverse than Imagen regardless of guidance; nevertheless, a study with additional factors other than linguistic features and random seeds could yield more insights.

Finally, it should be noted that our work intentionally disregards the relationship between quality and variability, although the two can be conflated.
For example, does increased variability reduce image quality?
Is Imagen a better option than, say, SDXL due to its higher quality, even if it generates less diverse imagery?
Thus, text-to-image models should not be chosen based on the findings of our study alone.
Rather, our work supplements image quality metrics in model selection.

\bibliography{anthology}

\appendix
\nobalance
\section{Detailed Experimental Settings}
\subsection{Computational Environment}\label{appendix:sec:development-environment}
Our primary software toolkits included HuggingFace Diffusers 0.25.0, Transformers 4.40.1, PyTorch 2.1.2, DreamSim 0.1.3, and CUDA 12.2.
We ran all experiments on a machine with four Nvidia A6000 GPUs and an AMD Epyc Milan CPU.

\begin{figure}[t!]
    \includegraphics[width=\linewidth,trim={0cm 0cm 5cm 0cm},clip]{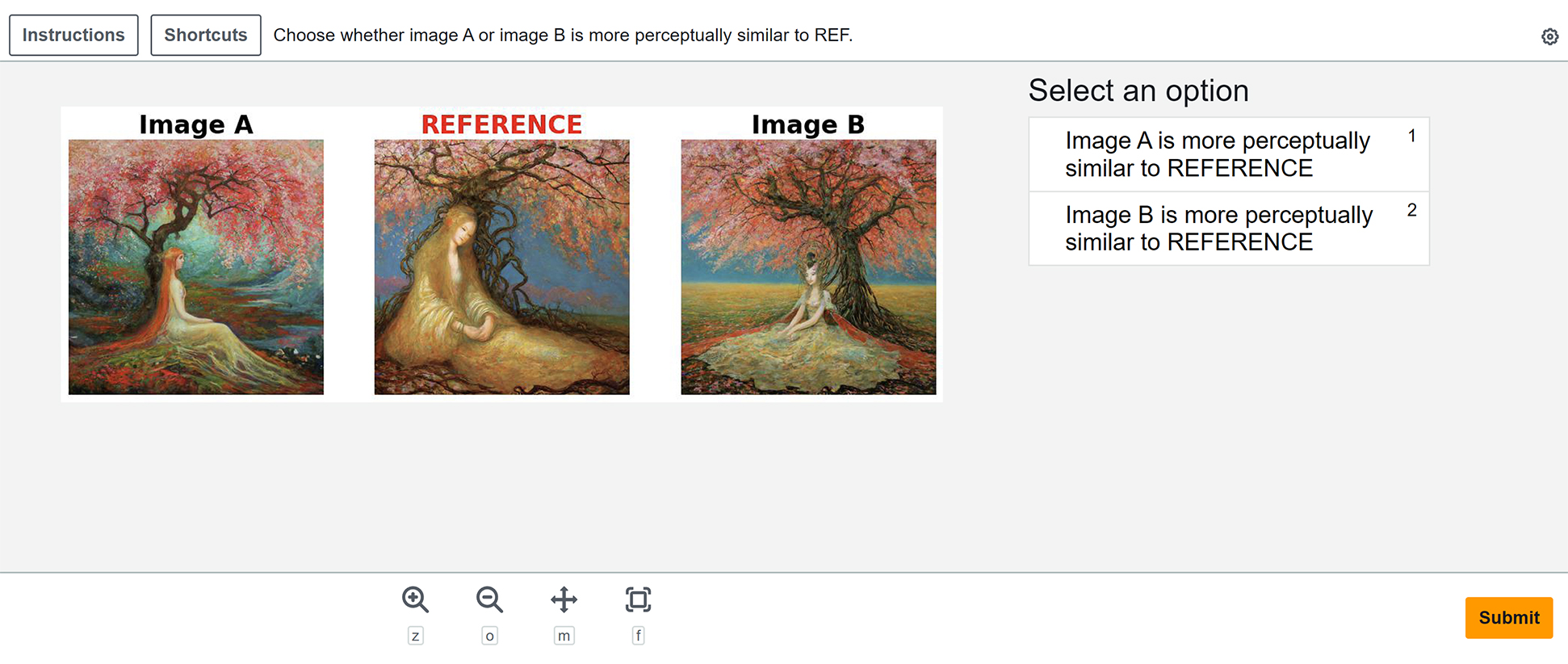}
    \caption{Interface for collecting 2AFC judgements.}\label{fig:2afc-ex}
\end{figure}

\subsection{Diffusion Model Details}\label{appendix:sec:diffusion-model-details}
\parheaderfirst{SDXL}
We downloaded \texttt{stabilityai/stable-} \texttt{diffusion-xl-base-1.0} from HuggingFace zoo.
We used the default guidance scale of 7.5 and 30 inference steps without the additional refiner module.
Each 1024x1024 SDXL image took 4--5 seconds to generate per card, resulting in a throughput of roughly 50--60 images per minute.

\parheader{Imagen}
We selected the \texttt{imagegeneration@006} model, the latest version as of April 2024, and generated four square images per call while varying the random seed.
This matched our SDXL throughput of 50--60 images per minute.
Each image was 1536x1536 in resolution.

\parheader{DALL-E 3}
For DALL-E 3, we used the default parameters of ``hd'' resolution (1024x1024) and ``vivid'' style.
To mitigate prompt editing, we followed the official documentation and prepended ``I NEED to test how the tool works with extremely simple prompts. DO NOT add any detail, just use it AS-IS: '' to the prompt.
The generation speed of DALL-E 3 was considerably slower than Imagen and SDXL at approximately 10 images per minute.

\subsection{Annotation Apparatuses}\label{appendix:sec:annotation-apparatuses}
We are deemed exempt by the \texttt{<blinded>} board of ethics requirements for review.

\parheader{W1KP quality}
We present the annotation user interface for collecting 2AFC judgements in \autoref{fig:2afc-ex}.
For our attention checks, we showed each worker at least one triplet with either image A or B exactly matching the reference.
If the correct answer was not chosen, we rejected all their labels and blocked them.
This resulted in a pass rate of around 90\%.
For higher quality, we required our workers to be ``Masters'' for participation eligibility.

\begin{figure}[t!]
    \includegraphics[width=\linewidth,trim={0cm 0cm 0cm 0cm},clip]{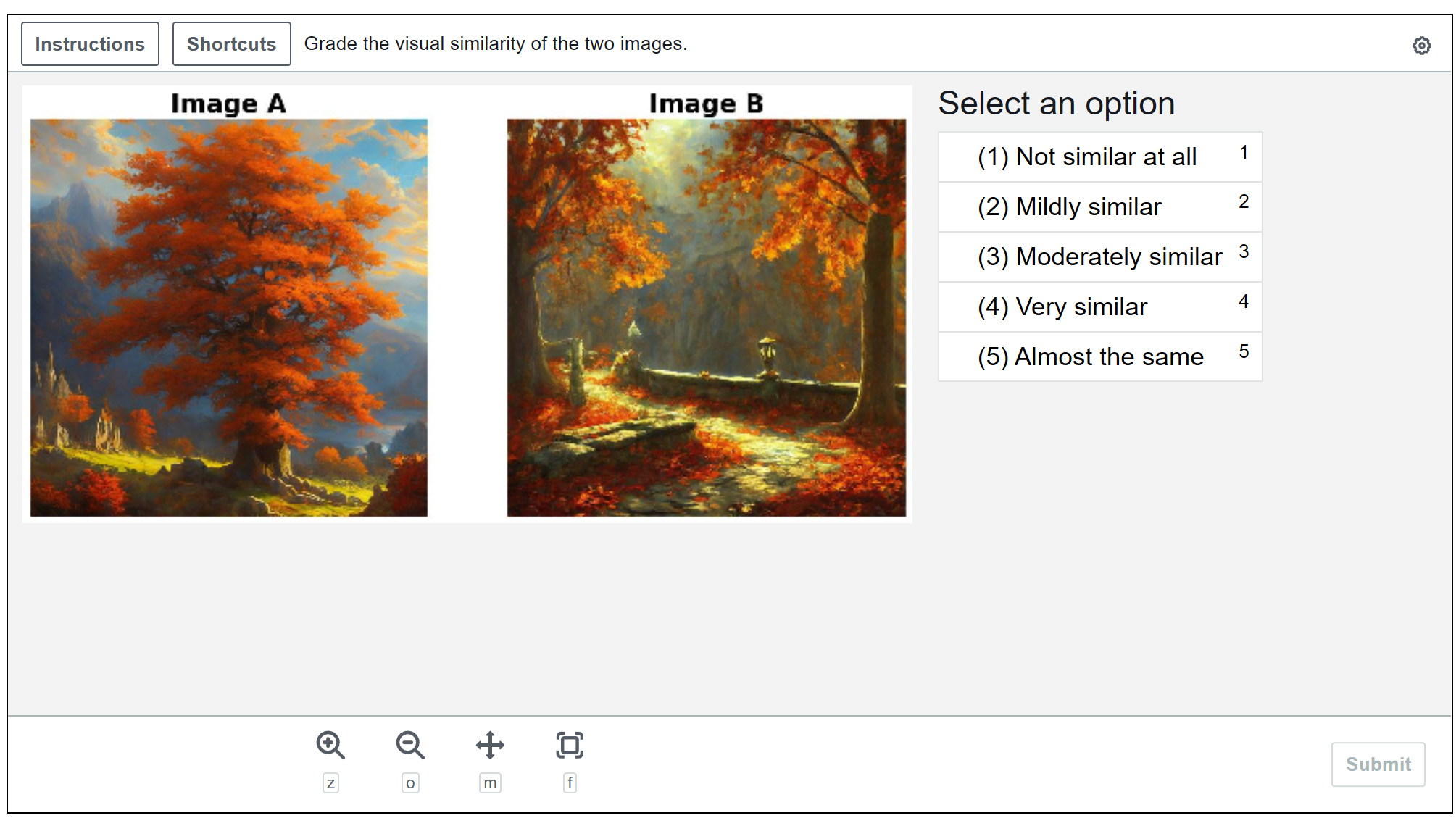}
    \caption{Interface for collecting graded judgements.}\label{fig:mi-ex}
\end{figure}

\parheader{W1KP metric interpretation}
We present our annotation interface for gathering graded similarity judgements in \autoref{fig:mi-ex}.
For the attention checks, we showed each annotator at least one pair of images that were the exact same.
If they did not choose ``almost the same,'' we discarded all their judgements, resulting in an acceptance rate of 95\%.

\section{Detailed Proofs}\label{appendix:sec:proofs}
\textbf{Proposition 2.1.}\textit{
    If $X$ is a continuous random variable, $F(X)$ is standard uniform $U[0, 1]$.
}

\begin{proof}
    Let $X$ be a continuous r.v. If $X$ is $U[0,1]$, then its CDF $\mathbb{P}(X \leq x) = x$. Since $\mathbb{P}(F(X) \leq x) = \mathbb{P}(X \leq F^{-1}(x)) = F(F^{-1}(x)) = x$, then $F(X)$ is $U[0,1]$, completing our proof.
\end{proof}

\noindent\textbf{Proposition 2.2.}\textit{
    If $d$ is the squared Euclidean distance and $h$ the pairwise mean kernel, $U_{d,h}$ is proportional to the trace of the covariance matrix of $f(I_1), \dots, f(I_n)$, i.e., the total variance.
}

\begin{proof}
    Consider the pairwise sum squared Euclidean distance $\sum_{i\neq j} || f(I_i) - f(I_j) ||^2_2$, which expands into
    \begin{equation}\small
        \sum_{i\neq j} f(I_i)^\intercal f(I_i) - 2f(I_i)^\intercal f(I_j) + f(I_j)^\intercal f(I_j).
    \end{equation}
    The first and third self-product terms expands as
    \begin{equation}\small
        (n-1)\sum_{i=1}^n f(I_i)^\intercal f(I_i)
    \end{equation}
    and
    \begin{equation}\small
        (n-1)\sum_{j=1}^n f(I_j)^\intercal f(I_j),
    \end{equation}
    and the middle term
    \begin{equation}\small
        \sum_{i,j}f(I_i)^\intercal f(I_j) - \sum_{i=1}^n f(I_i)^\intercal f(I_i).
    \end{equation}
    After algebraic manipulation, we arrive at
    \begin{equation}\small
        (n-1)\left(\frac{1}{n} \sum_{i=1}^n f(I_i)^\intercal f(I_i) - \frac{1}{n^2}\sum_{i,j} f(I_i)^\intercal f(I_j)\right).
    \end{equation}
    We are ready to relate this quantity to the trace of the covariance matrix, given by
    \begin{equation}\small
        \text{tr}(\Lambda) = \frac{1}{n}\sum_{i=1}^n || f(I_i) - \frac{1}{n}\sum_{j=1}^nf(I_j)||_2^2,
    \end{equation}
    which simplifies as
    \begin{equation}\small
        \frac{1}{n}\left(\sum_{i=1}^n f(I_i)^\intercal f(I_i) - \frac{1}{n}\sum_{i,j} f(I_i)^\intercal f(I_j)\right).
    \end{equation}
    Multiplying by $(n-1)$, we arrive at the sum of the pairwise squared Euclidean distance. Dividing by $n (n-1)$ yields the mean pairwise squared distance, and our proof is finished.
\end{proof}

\section{DiffusionDB Statistics}\label{appendix:sec:diffusiondb}
We now characterize the prompts and keywords in DiffusionDB.
To extract trailing keywords, we split prompts into a main part and its keywords part by applying these steps:
\begin{enumerate}[leftmargin=6mm,itemsep=0.25mm]
    \item Tokenize the prompt by commas, e.g., ``cat walking, 4k'' becomes ``cat walking'' and ``4k.''
    \item If any ``token'' after the first is shorter than four words, everything after that token is considered a keyword.
    \item The first ``token'' is always the main prompt.
\end{enumerate}
A preliminary analysis showed that this was more than 95\% accurate in identifying keywords.
We present ten examples below:
\begin{enumerate}[leftmargin=6mm,itemsep=0.25mm]
    \item ``ashtray in the messy desk of the detective, \underline{smoke and dark}, \underline{digital art}''
    \item ``onion very sad crying big tears cartoon, \underline{3d render}''
    \item ``the lost city of Atlantis, \underline{4K}, \underline{hyper detailed}''
    \item ``a galleon ship by Darek Zabrocki''
    \item ``hill overlooking a viking city, \underline{fantasy}, \underline{forested}, \underline{large trees}, \underline{top down perspective}, [...]''
    \item ``photo of an awesome sunny day environment concept art on a cliff, architecture by daniel libeskind with village, \underline{residential area}, \underline{mixed development}, \underline{highrise made up staircases}, [...]''
    \item ``giant oversized  battle hedgehog with army pilot uniform and hedgehog babies ,in deep forest hungle , \underline{full body} , \underline{Cinematic focus}, \underline{Polaroid photo}, \underline{vintage} , \underline{neutral dull colors}, \underline{soft lights}, [...]''
    \item ``pizza the hut, \underline{akira}, \underline{gorillaz}, \underline{poster}, \underline{high quality}''
    \item ``tengu spotted in atlanta''
    \item ``underground cinema, \underline{realistic architecture}, \underline{colorfull lights}, \underline{octane render}, \underline{4k}, \underline{8k}''
\end{enumerate}

\section{Visuolinguistic Analysis Details}
\subsection{Linguistic Feature Extraction}\label{appendix:sec:linguistic-feature-extraction}
For word sense clustering, we used the ``WN 2.1 -19370 synsets'' resource from \url{https://ai.stanford.edu/~rion/swn/}, previously published in \citet{snow2007learning}.
Unless otherwise stated, all our CLIP models were initialized from the \texttt{openai/clip-vit-large-patch14-336} checkpoint from HuggingFace, released by \mbox{OpenAI}.
Our GloVe embeddings were the 300-dimensional embeddings trained on 840B tokens of web text.

\subsection{Effects of Classifier-Free Guidance}\label{appendix:sec:cfg}
We briefly confirmed that increasing classifier-free guidance did not worsen the perceptual variability of SDXL below that of Imagen.
Imagen and DALL-E 3 do not expose classifier-free guidance as an input parameter, hence limiting us to SDXL.
We increased the classifier-free guidance from 5.0 to 30, much higher than the normal range of 5.0--7.5, and regenerated the images in Section~\ref{sec:visuolinguistic:prompt-reusability}.
We arrived at a mean W1KP score of 0.53 for SDXL, which was below Imagen's score of 0.62, e.g., SDXL still had greater variability.

\clearpage
\section{Dimensionality-Reducing Visualization}\label{appendix:sec:mds}

\begin{figure}[h!]
\centering
    \includegraphics[width=0.90\columnwidth]{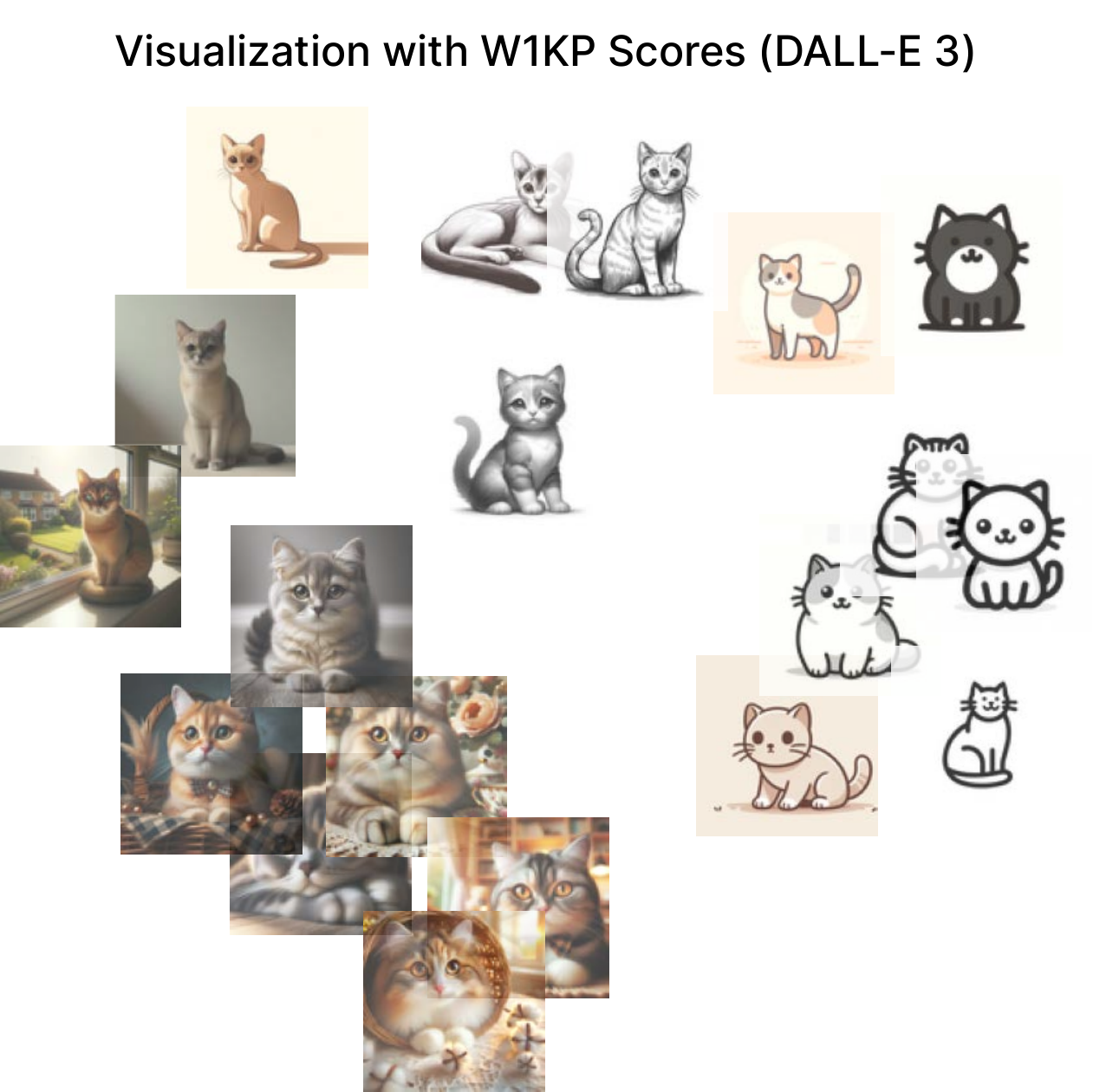}
    \noindent\rule{\columnwidth}{0.4pt}
    \includegraphics[width=0.90\columnwidth]{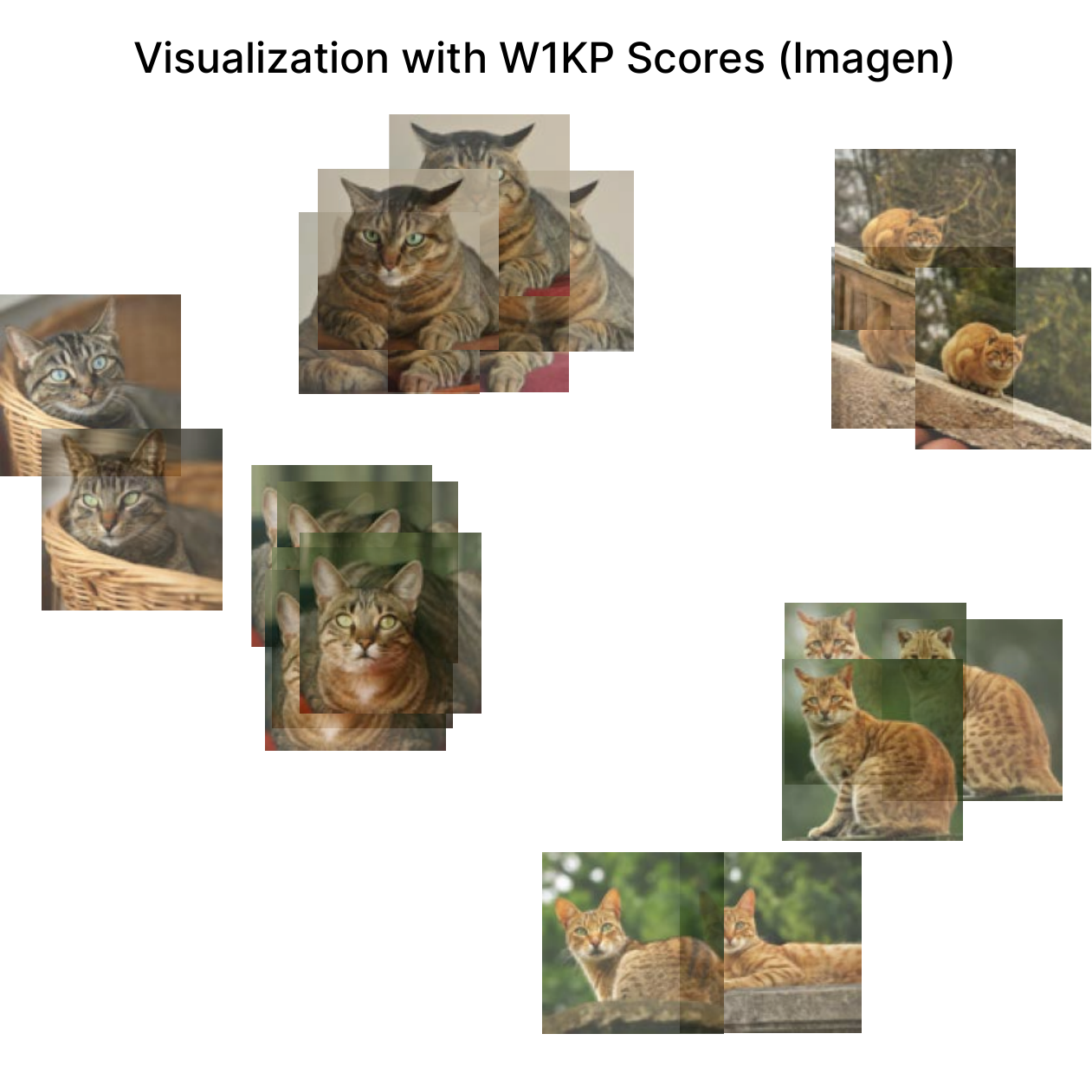}
    \noindent\rule{\columnwidth}{0.4pt}
    \includegraphics[width=0.90\columnwidth]{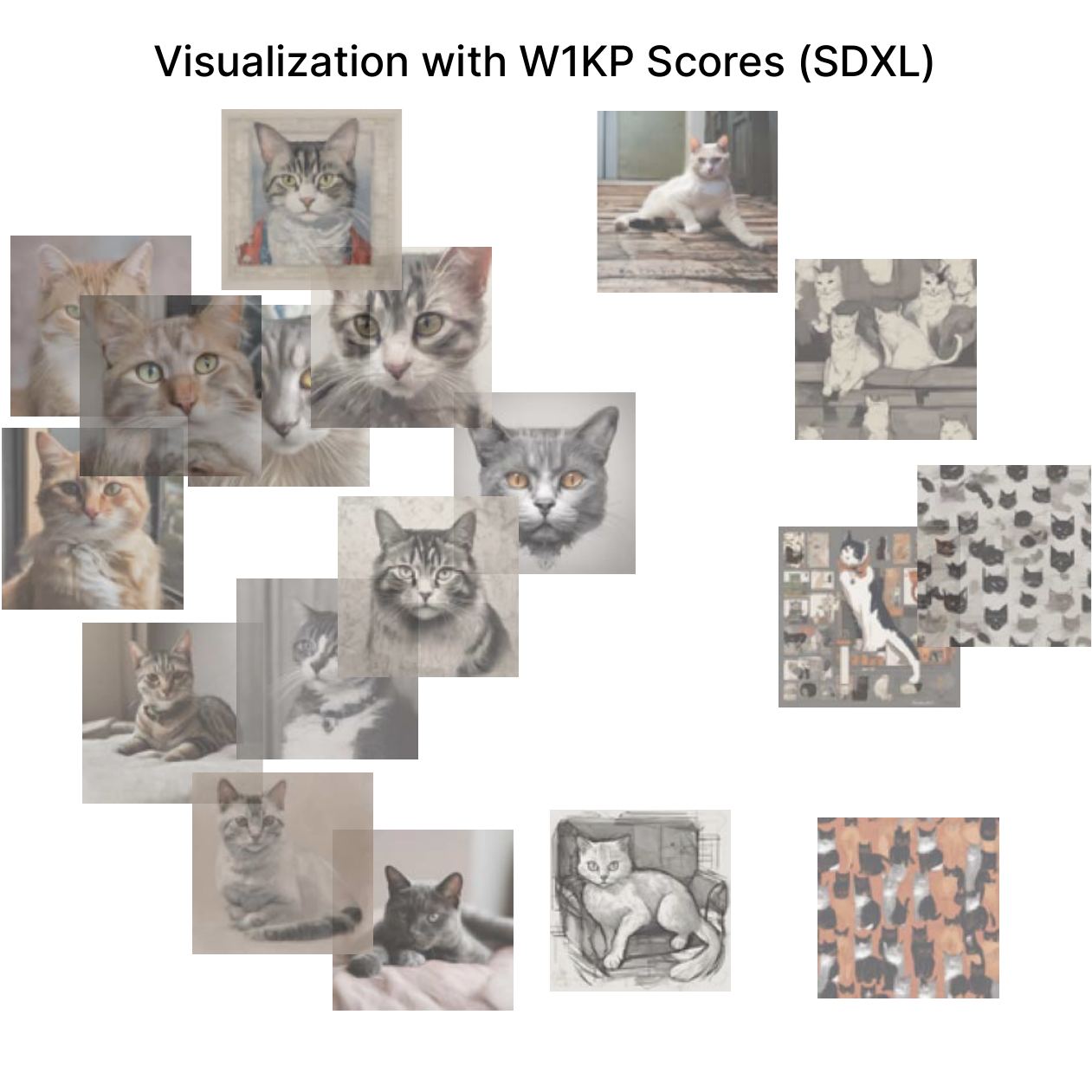}
    \caption{Twenty generated images for the prompt ``cat,'' clustered using multidimensional scaling on DreamSim$_{\ell_2}$. Imagen produces six distinct clusters.}
\end{figure}

\begin{figure}[h!]
\centering
    \vspace{1.32cm}
    \includegraphics[width=0.90\columnwidth]{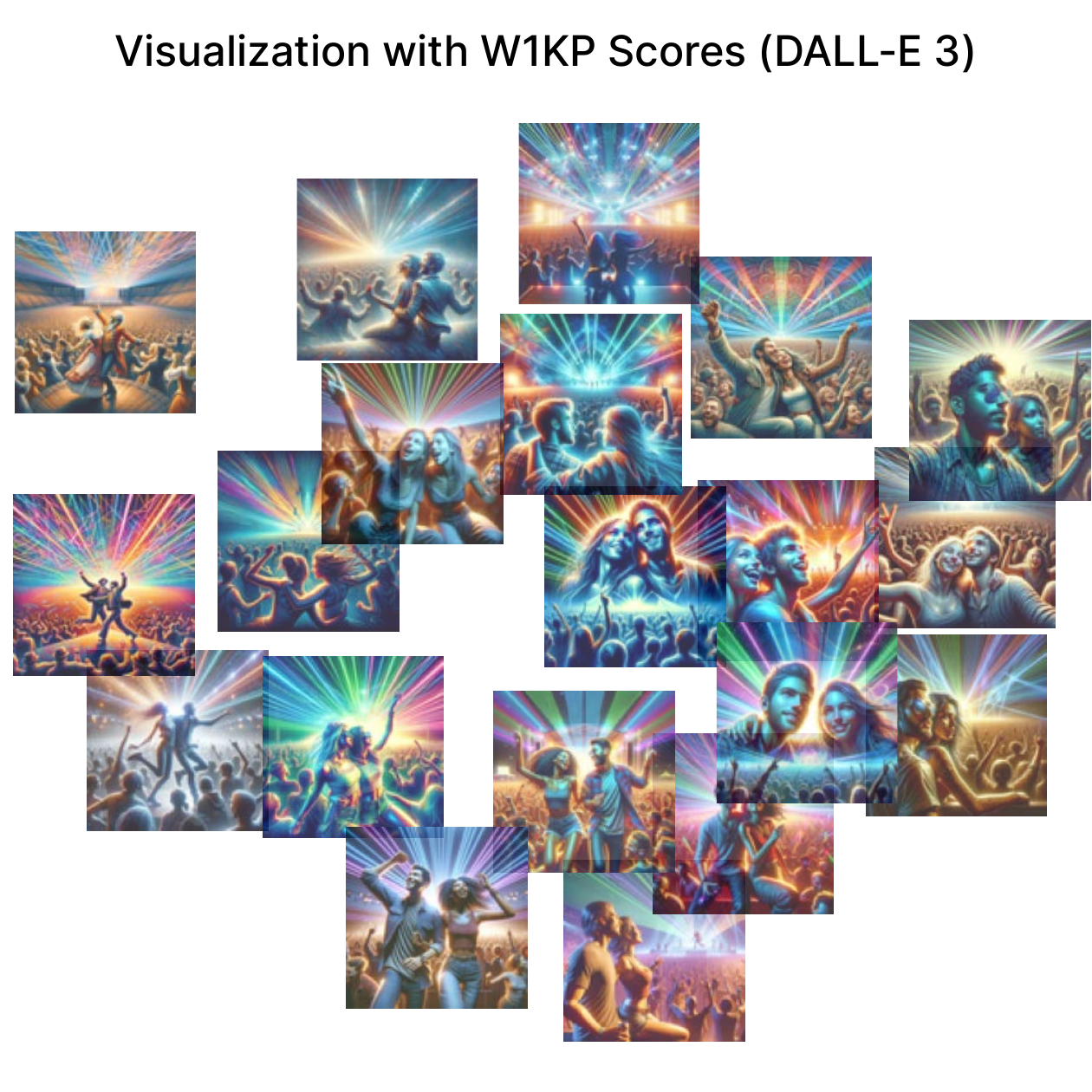}
    \noindent\rule{\columnwidth}{0.4pt}
    \includegraphics[width=0.90\columnwidth]{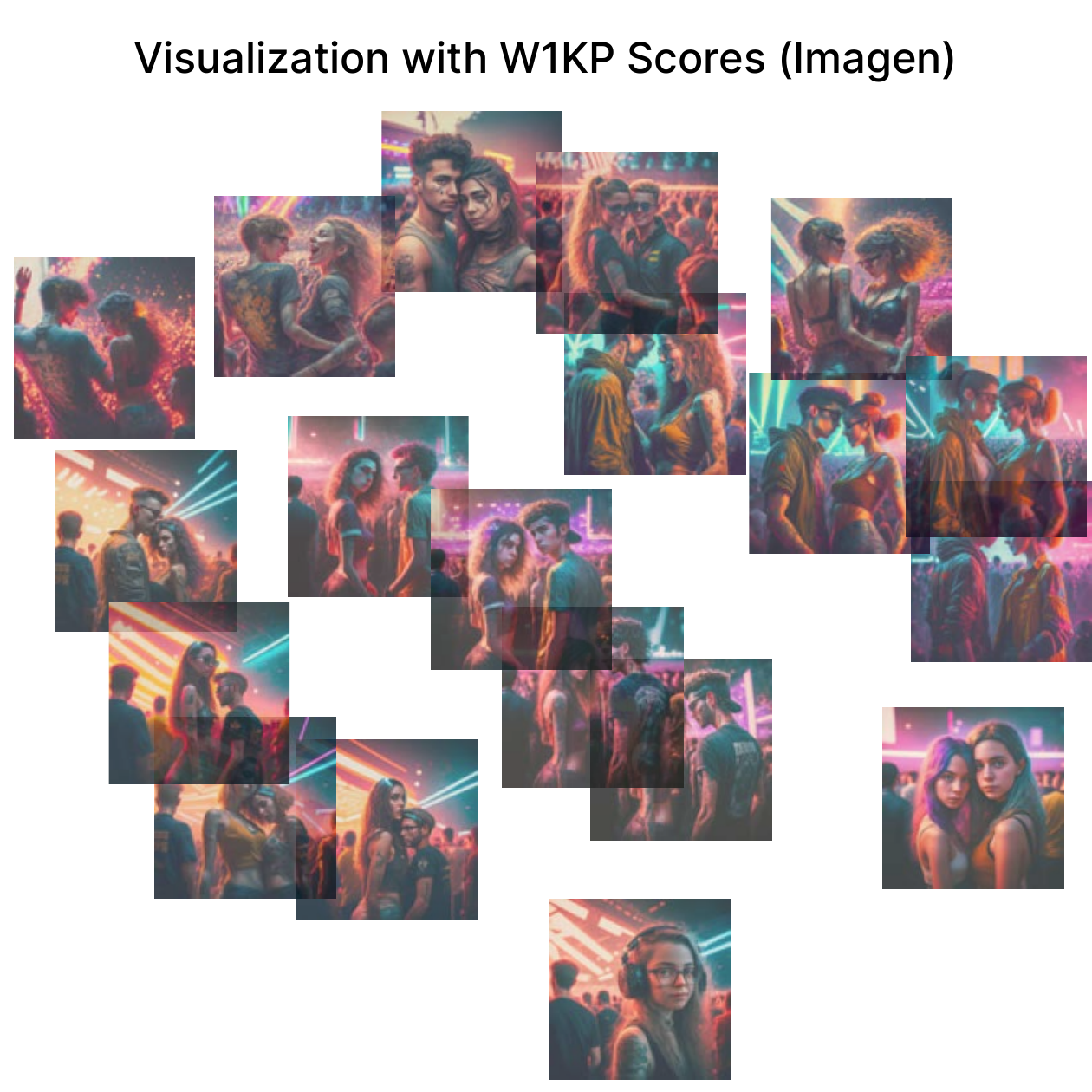}
    \noindent\rule{\columnwidth}{0.4pt}
    \includegraphics[width=0.90\columnwidth]{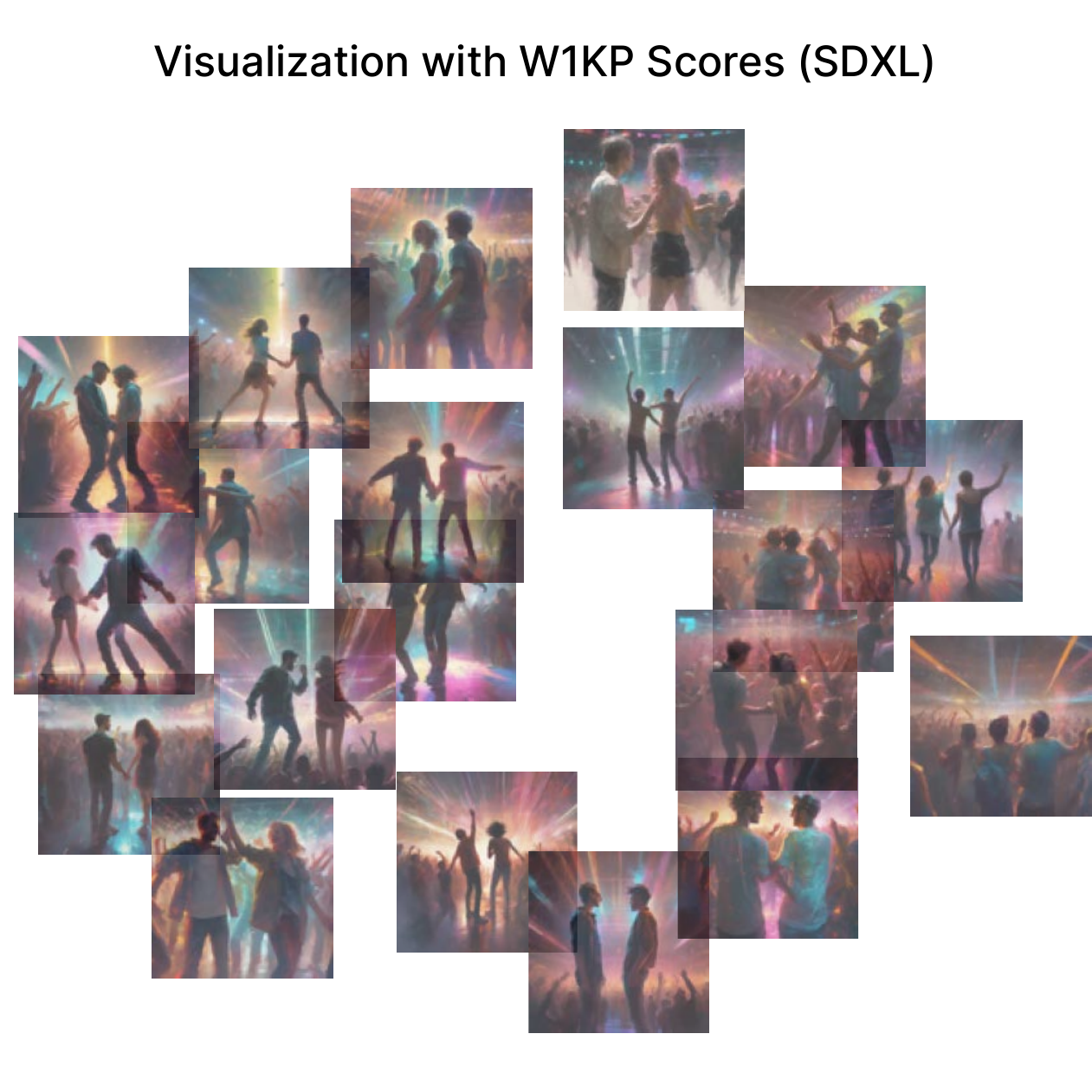}
    \caption{Generated images for a longer prompt.}
\end{figure}

\section{Supplementary Results and Discussion}

During peer review, our reviewers provided helpful feedback on the paper.
We explicitly address a few of their points below for transparency.

First, it was mentioned that reducing dissimilarity to a single numerical score does not do justice to all the nuances of image perception.
To this, we concur.
Summarizing a range of phenomena as a single scalar is a key drawback of any evaluation metric, and our approach is not different in this regard from well-established ones such as CLIP, BLEU, BERT score, Spearman's rho, Cohen's kappa, and others.
For example, a high BERT score or BLEU may not mean that translation quality is definitively good.
That remains to be judged on a task-by-task basis.

A second point from the reviewers was that our computational contribution in the current work was unclear, as our DreamSim model is only marginally better.
In our response, we emphasized that our key contributions are to propose and validate a human-calibrated framework for building variability metrics from existing baselines such as DreamSim-L2.
We examine a new practical application of the method and provide new linguistic insight.

A third question was about how a variability measure should balance between coverage and uniqueness, and how our measure supports this.
Such nuances are important to the design of the kernel function, for which we construct and analyze two chosen measures.
In the first pairwise-mean kernel ($\eta_\text{mean}$), all-pair similarities are weighted equally in a set.
Intuitively, this should provide a balanced assessment of overall variability (e.g., coverage), as every image pair has equal weight.
In the second $k$-expected maximum kernel ($\eta_k$), we estimate the maximum expected image-pair similarity out of a set of size $k$, thus focusing on the nearest pair of images (intuitively, the lack of uniqueness, e.g., duplicates in a set of size $k$).
Our choice of W1KP is further grounded by our human alignment, which provides interpretation of the scores.

Lastly, a few comments centered on the practical utility of obtaining multiple images from the same prompt.
In the multimedia industry, visual artists are tasked with storyboarding and brainstorming, which require creating different images of the same idea.
Our approach would assess the reusability of each prompt for that purpose before a prompt is considered ``used up.''

\subsection{Metric Interpretation Quantitative Study}
\label{appendix:sec:metric-interpretation}
One of the reviewers suggested quantifying the extent to which our W1KP cutoffs corresponded to qualitative features such as composition and style similarity, as claimed in Section~\ref{sec:veracity-analyses:interpretation}.
For this, we annotated 50 pairs of images, each from a different prompt from DiffusionDB, for each model.
For each image pair, we noted whether the two images matched in low-level features, high-level composition, and artistic style.
We found the following medians across the models:
\begin{table}[h!]\small
\setlength{\tabcolsep}{2.5pt}
\centering
\begin{tabular}{lccc}
\toprule[1pt]
\textbf{Rating} & \textbf{Feature Sim.} & \textbf{Composition Sim.} & \textbf{Style Sim.} \\
\midrule[1pt]
None & 18\% & 26\% & 24\% \\
Low & 24\% & 40\% & \textbf{62\%} \\
Medium & 66\% & \textbf{88\%} & \textbf{94\%} \\
High & \textbf{82\%} & \textbf{90\%} & \textbf{100\%} \\
\bottomrule[1pt]
\end{tabular}
\caption{The percentage of pairs matching in features, composition, and style, grouped by W1KP rating.}
\end{table}

The qualitative similarity increases with the rating, in order from low-level feature similarity to high-level style similarity, supporting our qualitative findings in Section~\ref{sec:veracity-analyses:interpretation}.

\end{document}